%% file: paper.tex
\documentclass[11pt, table]{article}

\usepackage[final]{acl}

\usepackage{times}
\usepackage{latexsym}

\usepackage[T1]{fontenc}
\usepackage{microtype}

\usepackage{inconsolata}

\usepackage{graphicx}

\usepackage{amsmath}
\usepackage{amssymb}
\usepackage{booktabs}
\usepackage{multirow}
\usepackage{cleveref}
\usepackage{float}
\usepackage{makecell}
\usepackage{threeparttable}
\usepackage{hyperref}

\newcommand{\fq}{\text{\#}}

\graphicspath{{res/}}
\title{Reducing Tokenization Premiums for Low-Resource Languages}

\author{
Geoffrey Churchill \and Steven Skiena \\
Stony Brook University
}

\begin{document}
\maketitle
\begin{abstract}
Relative to English, low-resource languages suffer from substantial tokenization premiums in modern LMs, meaning that it generally requires several times as many tokens to encode a sentence in a low-resource language than to encode the analogous sentence in English.
This tokenization premium results in increased API and energy costs and reduced effective context windows for these languages.
In this paper we analyze the tokenizers of ten popular LMs to better understand their designs and per-language tokenization premiums.
We also propose a mechanism to reduce tokenization premiums in pre-trained models, by post-hoc additions to the token vocabulary that coalesce multi-token characters into single tokens.
We apply this methodology to 12 low-resource languages, demonstrating that the original and compressed inputs often have similar last hidden states when run through the Llama 3.2 1B model.
\end{abstract}

\section{Introduction}

The very first step in the creation of any language model is the definition of its vocabulary, the set of atomic text blocks (tokens) into which input strings are segmented.
Most modern LMs have vocabularies of hundreds of thousands of tokens, typically including every distinct base character or byte pattern to ensure that every string can be encoded.
They also include longer strings representing common words and subwords, to compactly represent typical input strings.
Early tokenization strategies included breaking text at whitespace and punctuation, and more sophisticated handcrafted rules for splitting words into subwords.

Today, popular tokenizers derive their token vocabularies through data-driven approaches, typically {\em byte-pair encoding (BPE)} or a variant thereof.
BPE learns a tokenizer from a training corpus by starting with an initial set of possible characters, and then repeatedly merging the most common pair of tokens into a new token that is added to its growing vocabulary \citep{gage1994}.
This procedure continues for a given number of rounds until a desired vocabulary size is reached or, for smaller corpora, when the most common pair falls below some frequency threshold.

As a data-driven procedure, BPE learns a vocabulary that reflects the training corpus.
High-resource languages, especially English, typically dominate available training data, so the vocabularies of BPE-trained tokenizers largely comprise words and subwords from these languages. 
This has several undesirable consequences, as the tokenization premium (penalty) for important but lower-resource languages such as Bangla, Hindi, and Urdu can easily be three to five times that of English.
This penalty manifests itself in several hard costs and limitations: content in these languages will be processed slower, cost three to five times more in API calls/energy usage, and be limited to context windows of only a fraction of that available to English \citep{Petrov2023Unfairness}. 

In this paper, we study the landscape of current tokenizers to better understand their similarities and differences, and the tokenization premiums they impose on various languages.

We additionally explore an approach to mitigating tokenization premiums, by retrofitting a frozen model with additional tokens and derived (as opposed to trained) embeddings for these new tokens. Our approach presupposes that the model's training corpus included content in the target language, so that our challenge is to identify a single embedding for a common substring in the target language that the frozen model treats similarly to the substring's original multi-embedding encoding. This could yield more efficient tokenizations with hopefully only minor loss in quality.

Our main contributions in this paper are:

\begin{itemize}
    \item {\em Comparison of Vocabularies Across Models} -- Our analysis of the vocabularies of ten prominent LMs demonstrates that they cluster into several distinct groups according to Jaccard similarity.

    \item {\em Language-Specific Tokenization Penalties by Model} -- We show that models differ substantially in how efficiently they tokenize text between Latin and non-Latin low-resource languages. Important languages in non-Latin scripts (e.g. Bangla, Hindi, and Urdu) have tokenization penalties between three and four times that of English, while much smaller languages in Latin scripts (e.g. Danish, Swedish, and Norwegian) are efficiently encoded.

    \item {\em Mitigation Strategies for Language-Specific Tokenization Penalties} --
    We propose and evaluate different approaches to retrofitting a frozen model with new tokens to reduce tokenization premiums while minimizing the reduction in inference quality of the generated sequences. We focus our efforts on adding tokens to represent individual missing Unicode characters from non-Latin scripts.
\end{itemize}

Our paper is organized as follows.
We review previous work on tokenization strategies in \Cref{sec:related-work}.
We perform a comparative vocabulary analysis of ten popular tokenizers in \Cref{sec:comparing-tokenizers}, and report on language-specific tokenization premiums.
We propose and evaluate mitigation strategies for these languages in \Cref{sec:mitigation}, before concluding with possible future research directions in \Cref{sec:conclusions}.

\section{Related Work}
\label{sec:related-work}

We consider related work in (i) tokenization strategies and (ii) post-training adaptations of frozen models.

\paragraph{Tokenization:}
The most popular tokenization strategies used by LMs are based on byte-pair encoding (BPE). This strategy was first proposed by \citet{gage1994} in the context of dictionary-based compression. The idea is simple: initialize the vocabulary of tokens to the set of all characters, and in each iteration merge a most frequent adjacent pair of tokens in the corpus into a single new token which is added to the growing vocabulary. Running BPE on \texttt{"she\_shakes\_shoes"} until all pairs are singletons would (up to tie-breaking) look like:

\begin{table}[H]
    \tiny\begin{tabular}{c|l|c}
        Tokenized string & Vocabulary & Merge \\
        \hline
        \texttt{s h e \_ s h a k e s \_ s h o e s} & \texttt{\_ a e h k o s} & \texttt{s h} \\
        \texttt{sh e \_ sh a k e s \_ sh o e s} & \texttt{\_ a e h k o s sh} & \texttt{\_ sh} \\
        \texttt{sh e \_sh a k e s \_sh o e s} & \texttt{\_ a e h k o s sh \_sh} & \texttt{e s} \\
        \texttt{sh e \_sh a k es \_sh o es} & \texttt{\_ a e h k o s sh \_sh es} &  \\
    \end{tabular}
\end{table}

\citet{sennrich2016bpe} applied BPE to improving the machine translation of rare words. BPE improves upon naive methods like whitespace splitting (which yields a huge set of opaque tokens), or handcrafting a more morpheme-aware tokenizer (which is brittle and does not scale). Whitespace splitting would yield unrelated tokens for \texttt{"unrelated"} and \texttt{"relatable"} whereas BPE could tokenize them as \texttt{"un|relat|ed"} and \texttt{"relat|able"}, which share the \texttt{"relat"} subword and contain the informative morphemes \texttt{"un"}, \texttt{"ed"}, and \texttt{"able"}.

\citet{Petrov2023Unfairness} and \citet{ahia2023languagescostsametokenization} explore and quantify some of the downstream impacts of imbalanced BPE training corpora, showing that applying LMs to content in low-resource languages is often much slower and many times more costly than the analogous English content.

WordPiece \citep{schuster2012wordpiece} is a related strategy that, like BPE, iteratively merges pairs of tokens. WordPiece, however, is explicitly geared towards language modeling by merging the pair that would most increase the likelihood of the training corpus, with respect to some language model. For example, according to a unigram model, the log likelihood of the training corpus $C$ is
\[
\sum_{x \in C} \log p(x) = \sum_{x \in C} \log \fq x - |C| \log |C|
\]
where $x \in C$ ranges over every token in $C$, and $\fq x$ denotes the frequency of $x$ in $C$.
Merging every $\texttt{a b}$ pair into an $\texttt{ab}$ gives a net reduction of $\fq \texttt{ab}$ tokens, where $\fq\texttt{ab}$ is computed ``sequentially'', i.e. without overlaps, since otherwise if $\texttt{a}$ and $\texttt{b}$ were the same token the bigram frequency would be overcounted whenever the token occurred in a run of length > 2. Then the post-merge log likelihood is maximized by merging the $\texttt{a}$ and $\texttt{b}$ that maximize

{\footnotesize
$$
\left(\fq\texttt{ab}\right) \log\frac{\fq\texttt{ab}}{\left(\fq\texttt{a}\right) \left(\fq\texttt{b}\right)} - \left(|C| - \fq\texttt{ab}\right)\log\left(|C| - \fq\texttt{ab}\right)
$$
}%

The ``unigram language model'' (ULM) strategy \citep{kudo2018unigram} is essentially a top-down version of the bottom-up WordPiece. Instead of iteratively merging pairs of tokens, it removes tokens from a vocabulary that is initialized to some large seed vocabulary. As in WordPiece, the loss is based on the likelihood (explicitly according to a unigram model in the ULM paper), but ULM removes the token that least decreases the likelihood, rather than merging the pair that would most increase the likelihood. The ULM paper furthermore proposes a probabilistic approach to both loss calculation and decoding mechanisms that explores the most likely tokenizations of a string.

Recently, \citet{pagnoni2024bytelatenttransformerpatches} introduced the Byte Latent Transformer (BLT), which generalizes the fixed embedding matrix to a lightweight byte-level encoder/decoder pair whose job is to produce on-the-fly embeddings for dynamically segmented ``patches'' of bytes, where patch boundaries are drawn whenever a lightweight autoregressive model exceeds an entropy threshold (either global or relative to the entropy at the previous byte). This is an appealing idea that is more scalable than bottom-up vocabularies that contain a large number of tokens that are substrings of other tokens.

\paragraph{Frozen Model Adaptation:}
Given the high cost of retraining large language models, we consider approaches to add new tokens and associated embeddings to frozen models. Even when a model or embedding will be (re-)trained, such approaches may be useful as an initialization step.

The problem of adapting a frozen black box system to a task has been studied in many contexts. The adaptation problem addressed in the current paper is relatively easy in the sense that our frozen black box is at least already geared towards human language, whereas \citet{pang2024LM4VisualEncoding} shows that a frozen transformer layer from a unimodal {\em language} model can successfully encode {\em visual} tokens for various downstream tasks. Reservoir computing \citep{schrauwen2007reservoir} trains an encoder/decoder pair to adapt a fixed and often randomly generated dynamical system to a particular sequence processing task.

\section{Comparing Tokenizers}
\label{sec:comparing-tokenizers}

We evaluated ten different tokenizers from a range of models and release dates. \Cref{table:tokenizers} provides basic information about each tokenizer and a popular model that uses it.

There is a wide range in the vocabulary sizes (from 32K to 200K tokens) and release dates (from 2018 to 2024), capturing much of the important history of LMs.

\begin{table*}
\centering
\begin{threeparttable}
\footnotesize

\begin{tabular}{ccrrccc}
Company & Model & \makecell{Normalized\\vocab size} & Context size & Release date & Reference \\
\hline
Google & BERT\tnote{1} & 119,543 & 512 & Oct 2018 & \citet{bert}\\
OpenAI & GPT-2 & 50,256 & 1,024 & Feb 2019 & \citet{gpt2}\\
Google & T5 & 32,100 & 512 & Oct 2019 & \citet{t5}\\
Facebook & M2M-100 & 128,104 & 1,024 & Oct 2020 & \citet{m2m100}\\
OpenAI & GPT-3.5 & 100,256 & 16,385 & Mar 2022 & \citet{chatgpt}\\
Mistral AI & Mistral 7B & 32,515 & 4,096\tnote{2} & Sep 2023 & \citet{mistral7b} \\\
Anthropic & Claude 2.1 & 64,298 & 200,000 & Nov 2023 & \citet{claude21}\\
OpenAI & GPT-4o & 199,998 & 128,000 & May 2024 & \citet{gpt4o}\\
Meta & Llama 3 & 127,040 & 131,072 & Sep 2024 & \citet{llama3}\\
AllenAI & OLMo 2 & 99,579 & 4,096 & Dec 2024 & \citet{olmo20252olmo2furious}\\
\end{tabular}
\begin{tablenotes}
\item[1] This is actually BERT Multilingual, but is abbreviated to BERT in the tables and figures.
\item[2] Mistral 7B uses a sliding window mechanism that allows information to propagate beyond the immediate 4,096 token window.
\end{tablenotes}
\end{threeparttable}
\caption{Descriptive information on the ten tokenizers we evaluated, including a popular model using each.}
\label{table:tokenizers}
\end{table*}

\subsection{Vocabulary Comparisons}
\label{sec:vocabulary-comparison}

\begin{figure}
    \centering
    \includegraphics[width=\linewidth]{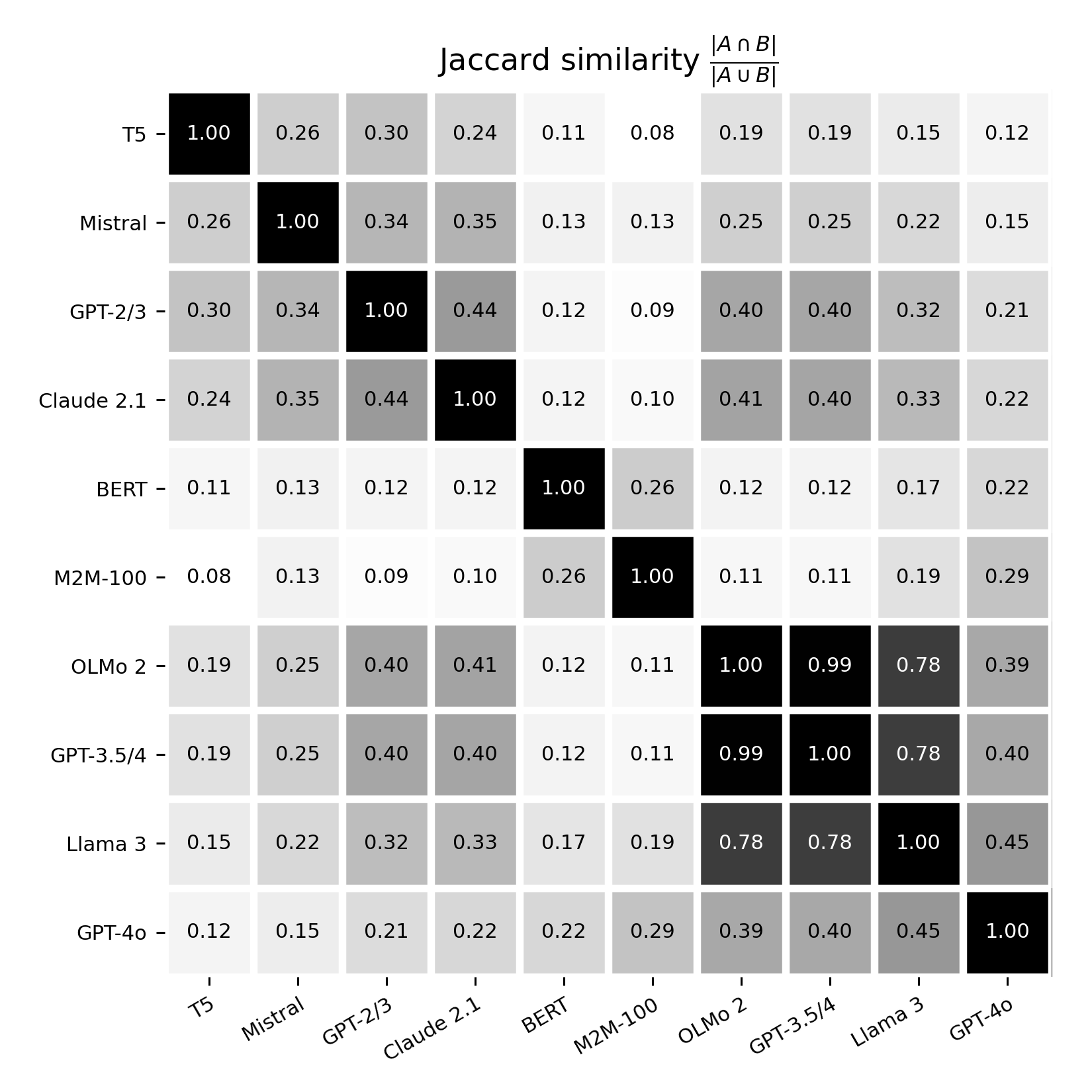}
    \caption{Jaccard similarities between normalized tokenizer vocabularies.}
    \label{fig:tokenizers/JACCARD}
\end{figure}

We compare tokenizer vocabularies in \Cref{fig:tokenizers/JACCARD} using Jaccard similarity. Tokens are compared for exact equality, so overlaps and substrings are not taken into account. Vocabularies are first normalized to account for tokenizer-specific markers like \texttt{"Ġ"} and \texttt{"\#\#"}, using the \href{https://github.com/huggingface/transformers/blob/dd16acb8a3e93b643aa374c9fb80749f5235c1a6/src/transformers/generation/stopping_criteria.py#L277}{\texttt{\textbf{clean\_tokenizer\_vocab}}} method from HuggingFace's \texttt{transformers} Python library. 

This figure highlights three clusters of tokenizer vocabularies, sorted roughly in order of increasing vocabulary size.

The first cluster comprises the four smallest vocabularies, from T5 (Google), Mistral (MistralAI), GPT-2 (OpenAI), and Claude 2.1 (Anthropic). T5 derived its vocabulary from the Common Crawl, which likely makes up a significant portion of the other three vocabularies in the cluster.

The second cluster comprises BERT Multilingual (Google) and M2M-100 (Meta), both ``multilingual'' models. The multilingual cluster presumably arises mostly by virtue of overlap in their target languages and training corpora -- the cluster is noticeable not because the two vocabularies are especially similar, but because they are so distinct from the other vocabularies.

The third cluster comprises GPT-3.5/4 (OpenAI), OLMo 2 (AllenAI), and LLama 3 (Meta). This is expected, as both OLMo 2 and Llama 3 base their vocabularies upon the GPT-3.5/4 vocabulary.

The last and largest vocabulary is that of GPT-4o (OpenAI). It contains 200K tokens and largely subsumes most of the other vocabularies (see \Cref{fig:tokenizers/PERCENT_SMALLER} in the appendix).

\Cref{table:tokenizer_vocab_breakdown} collects several statistics on each tokenizer vocabulary, again after normalization with \href{https://github.com/huggingface/transformers/blob/dd16acb8a3e93b643aa374c9fb80749f5235c1a6/src/transformers/generation/stopping_criteria.py#L277}{\texttt{\textbf{clean\_tokenizer\_vocab}}}.

\subsection{The FLORES-200 dataset}
\label{sec:flores}

The FLORES-200 dataset \citep{teamNoLanguageLeft2022} comprises curated human translations of each of 3001 sentences (averaging 21 words per sentence) into each of 200 written languages. Several languages have translations in two scripts, e.g.\ Mandarin Chinese (\texttt{zho}) in simplified (\texttt{Hans}) and traditional (\texttt{Hant}) scripts, Modern Standard Arabic (\texttt{arb}) in Arabic (\texttt{Arab}) and Latin (\texttt{Latn}) scripts, and Tamasheq (\texttt{taq}) in Tifinagh (\texttt{Tfng}) and Latin (\texttt{Latn}) scripts.

\subsection{Language-Specific Tokenization Premiums}

The ``dev'' split of the FLORES-200 corpus provides 997 sentence translation pairs between English and a given language $X$, for each of 200 distinct languages.
As the semantic content of both sentences in a pair is roughly equivalent, the number of tokens used to encode them for a given LM would ideally also be comparable.
Following \citet{Petrov2023Unfairness}, the {\em tokenization premium} for language $L$ is the average ratio of the number of tokens used to encode a sentence in language $L$ divided by the number of tokens used to encode the corresponding English sentence.

\begin{table*}
\footnotesize\input{res/FLORES-200/tgt_to_src_token_ratios_per_tokenizer_color}
\caption{Premiums for each tokenizer on various languages.}
\label{table:premiums_per_tokenizer}
\end{table*}


\Cref{table:premiums_per_tokenizer} shows per-tokenizer premiums for various languages. Shan (\texttt{shn}) is the most interesting language here, incurring a staggering 19x premium with the GPT-2/3 tokenizer, whereas with BERT Multilingual it is tokenized more than twice as efficiently as English. Chinese (\texttt{zho}) is tokenized five times more efficiently than English by T5, despite it having been trained on English, German, French, and Romanian, so it presumably incorporated many Chinese characters in the initial seed vocabulary.

\begin{figure}
    \centering
    \includegraphics[width=\linewidth]{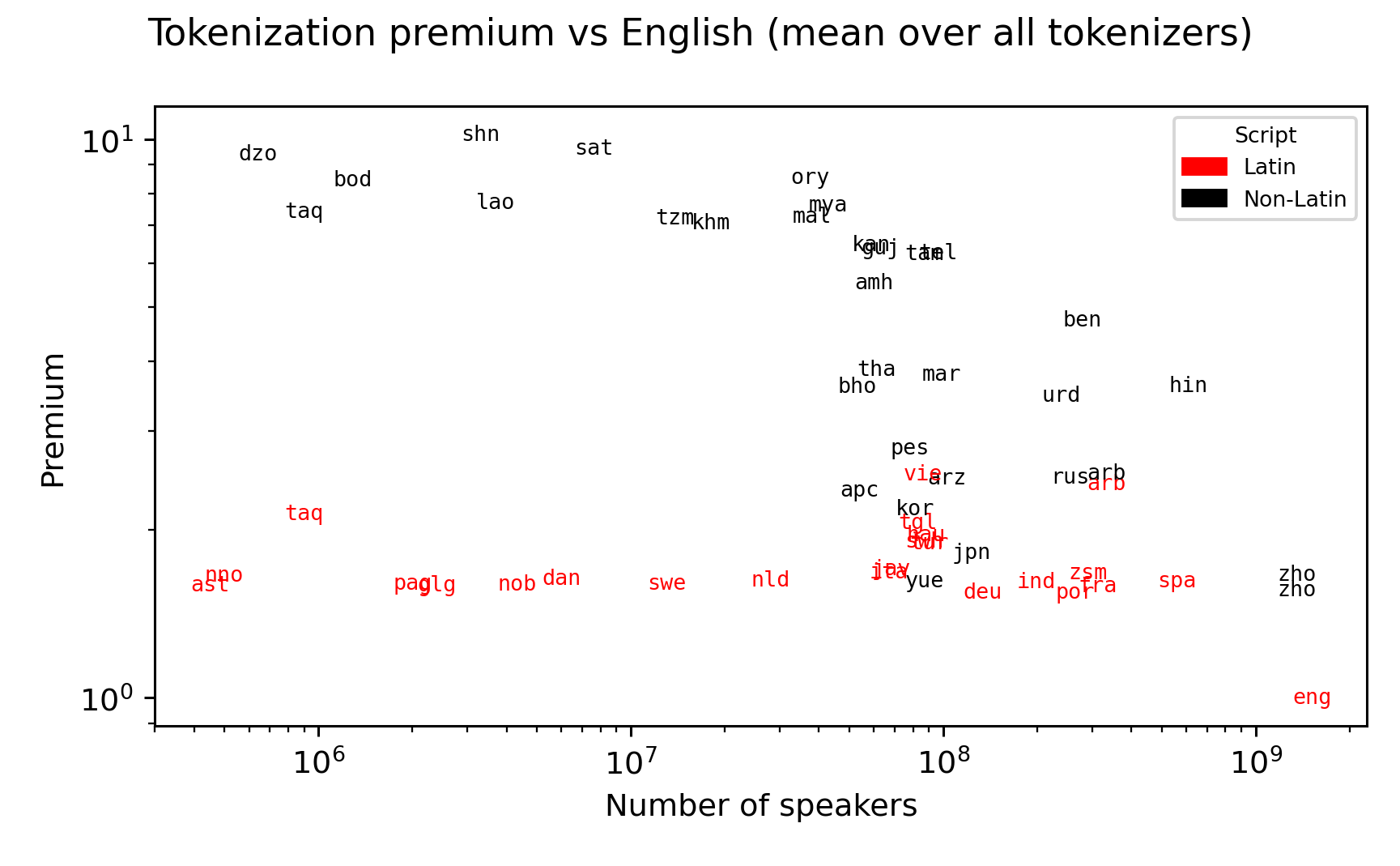}
    \caption{Tokenization premiums of several languages by number of speakers, averaged over all tokenizers. Some language codes, like "arb" and "zho", are available in two scripts and so appear twice in the plot.}
    \label{fig:flores/languages}
\end{figure}

\Cref{fig:flores/languages} plots several languages by estimated number of total speakers and tokenization premium. Moving leftward from the right, toward languages with fewer speakers, there is a clear fork into a lower branch where the Latin scripts remain very well-tokenized, and an upper branch where non-Latin scripts incur a steep penalty. Hindi (\texttt{hin}), Bangla (\texttt{ben}), and Urdu (\texttt{urd}) stand out for being both widely spoken and poorly tokenized, suggesting a distributional difference between what is spoken and what winds up in training corpora, possibly contributed to by people preferring English for creating internet content even when it is their second or third language, in order to reach a broader audience. Tamasheq (\texttt{taq}) shows the most drastic reduction in premium going from its native Tifinagh script to its Latin transliteration.

\begin{table*}
\tiny\input{res/tokenizers/tokenizer_vocab_breakdown}
\caption{Breakdown of cleaned vocabularies. Distinct Unicode blocks were counted by mapping every character in every token to its block. The block counts for OpenAI's vocabularies could in theory be off, because they provide their vocabularies in raw bytes, and multi-byte characters are sometimes split apart in the vocabulary. A best-effort attempt was made to recover characters: for every byte string, the longest valid UTF-8 substring (if one existed) contributed its characters to the total character set, subject to the constraints that no more than three bytes were removed from each of the front and back (UTF-8 uses up to four bytes per character), and that at least two bytes remained (this constraint is actually unnecessary as all 128 ASCII characters were already in every OpenAI vocab, and the other 128 bytes are not themselves valid UTF-8).}
\label{table:tokenizer_vocab_breakdown}
\end{table*}

\section{Mitigating Tokenization Premiums in Pre-Trained Models}
\label{sec:mitigation}

The best way to minimize language-specific tokenization penalties is careful up-front vocabulary selection, to include a sufficiently rich set of language-specific tokens before training a model.

Here we consider lightweight (training-free) methods of retrofitting a frozen model with additional tokens and untrained embeddings to reduce language tokenization penalties without significantly impacting model performance.

For this preliminary study, we focus on the special case of replacing multi-token encodings of OOV Unicode characters with new, per-character tokens. The primary technical issue is identifying an appropriate embedding for each new token so that it behaves similarly to the original multi-token encoding.

\begin{table*}
    \centering
    \scriptsize\input{res/llm_evaluation/compare_doc_embeddings_color}
    \caption{Comparison of various embedding strategies and languages. The cells are the cosine similarities of the averaged last hidden state with that of the original Llama 3.2 1B model output, averaged over the subcorpus of FLORES-200 corresponding to each language. \texttt{zho} denotes traditional Chinese script.}
    \label{table:doc_cosine_similarities}
\end{table*}

\subsection{Mitigation Methods}
\label{sec:mitigation-methods-subsection}

We experiment with several methods of deriving an embedding for a token newly added to a vocabulary. For each method $m$, we
\begin{enumerate}
    \item Use the original tokenizer to tokenize the string $s$ that will become the new token, yielding a list of embeddings $E_s$.
    \item Run the model (Llama 3.2 1B) on $E_s$ to some depth/layer $l$ with no special tokens (\texttt{<bos>}, \texttt{<pad>}, etc), yielding $|E_s|$ hidden states $H_s$. Layer 0 is the input embedding layer where $E_s=H_s$, layer 1 contains the outputs of the first transformer block, and so on until layer 16, which contains the outputs of the last transformer block, which are then multiplied by the transposed input embedding matrix to yield logits for the next token prediction.   
    \item Average $H_s$ into a single vector $h_s$. When $l=0$ then $h_s$ is just the average of the input embeddings $E_s$.
    \item Use $m$ to predict a single input space vector that, if run on its own to the same depth, would yield a similar $h_s$.
\end{enumerate}

To train our prediction models we make use of a per-layer matrix of hidden representations $V_l$, obtained by running the Llama model separately on each of its 128K vocabulary embeddings (from the input embedding matrix $V_0$) to layer $l$, so that each original token $t$ has a corresponding hidden representation in $V_l$. $V_0$ could in principle be augmented with any number of synthetic test embeddings.

We use the following prediction methods:

\paragraph{$k$-nearest neighbors:}
We build a nearest neighbors structure over the 128K hidden embeddings in $V_l$. To predict an input embedding for a string with averaged hidden representation $h$, we find the $k$ nearest neighbors to $h$ and average their input embeddings in $V_0$ weighted by their inverse distances from $h$, where the distances are with respect to the hidden representations in $V_l$.

\paragraph{Linear regression:}
We learn an affine transformation (linear plus bias) from $H_V$ to $E_V$ to predict an input embedding.

\paragraph{Local linear regression:}
This approach is a combination of the other two. To predict an input embedding for $h$, we find its $k$ nearest neighbors, train a weighted linear regression from those $k$ hidden representations to their $k$ corresponding input embeddings, and apply the learned transformation to $h$. In this case, the weight for distance $d$ is $\frac{1}{\exp d}$, instead of the $\frac{1}{d}$ we use in $k$-nearest neighbors.

\subsection{Experimental Results}
\label{sec:mitigation-experiments-subsection}

To evaluate the embedding strategies, we measure how similarly the Llama 3.2 1B model encodes sentences at its last hidden layer (layer 16) before and after vocabulary augmentation. In particular, for a given language we run the model on each sentence in that language both before and after augmentation. From each of the two runs we extract the last hidden state and average it into a single 2,048-dimensional vector. We take the cosine similarity between these two vectors to be a per-sentence similarity measure, which is averaged over all sentences to produce a similarity for a particular language and strategy. We limit the vocabulary augmentation to only include individual characters absent from the original vocabulary. This means that most Latin-script languages, as well as some other popular languages (e.g. Arabic) always have similarity of 1.0.

Table \ref{table:doc_cosine_similarities} reports, for each embedding strategies and several languages, these similarity scores. There are several interesting phenomena:

\begin{itemize}
    \item Linear regression at layer 0 is the best strategy on average. It is actually equivalent to simply averaging the input embeddings, as the learned transformation is the identity. It is easily the best strategy for Shan (\texttt{shn}), which is by far the worst tokenized language.
    \item At layer 0, Amharic (\texttt{amh}) has extremely low similarity using any of the local methods. 
    \item At layer 0, local linear regression is more robust than KNN, and tends to match or nearly match the best performing KNN.
    \item Vietnamese (\texttt{vie}) shows high similarities, but only 26\% of the tokens in the encoded sentences were new, so the model was less impacted by augmentation in the first place.
    \item Traditional Chinese script (\texttt{zho}) shows very high similarities, possibly due to it having dedicated Unicode blocks on account of its large alphabet, so the constituent tokens of a multi-token Chinese character may be sufficiently orthogonal to other tokens that the averaging operation is less lossy than in other languages.
    \item Unsurprisingly, linear regression performs worse in later layers.
    \item The worst languages for local strategies at layer 0 (\texttt{amh} and \texttt{shn}) are better preserved by the same strategies at later layers, suggesting that, as expected, the model determines appropriate contextualized representations for the constituents of multi-token characters.  
\end{itemize}

\section{Conclusions and Future Directions}
\label{sec:conclusions}

We have demonstrated that popular LMs impose substantial tokenization penalties on many widely spoken languages, and particularly on non-Latin scripts. These penalties are essentially a language-specific tax, manifesting as increased API costs, increased computational costs/electrical usage, and reduced effective input windows. We proposed an approach to mitigating this penalty through post-training vocabulary modification.

One dimension of future work is in vocabulary selection. BPE-like prioritize popular substrings, but their training corpora are largely based on internet content which is naturally skewed toward the most popular languages. On the other hand, when people interact with a chatbot, they probably prefer to use their first language. During vocabulary selection, tokens could be re-weighted to incorporate e.g. the estimated number of first-language speakers. Another option, that might incentivize cross-lingual transfer learning and make better use of the embedding space, could incorporate tag embeddings that explicitly mark the language(s) of the current context.

Another possible avenue of exploration is a sort of continuous/relaxed pseudo-translation of sequences: given a corpus $L$ of content in a low-resource language and a fixed, frozen language model $M$, learn a custom tokenizer and embedding matrix for $L$, and a mapping from sequences of $L$ embeddings to sequences of $M$ embeddings. While a true, high-quality, discrete token-to-token translation is a privilege enjoyed only by heavyweight neural models, the hope is that a simple model would suffice for pseudo-translation in the continuous embedding space. A good pseudo-translation, while unreadable to humans, might be readable to an LM presented with the raw embedding sequence, so if it is possible to create good pseudo-translations, they could be used as lightweight post-training adapters that can convert a user's preferred language into an LM's preferred language. If the LM's output is text, converting back into the user's language would require a heavyweight model, but if its output is an embedding, then the user might leave it as-is, or else translate it back via the reverse mapping.

It may be possible to further scale fixed vocabulary approaches, enabling more languages to be incorporated into vocabularies. One idea, inspired by fastText's averaging of subword embeddings \cite{bojanowskiEnrichingWordVectors2017a}, is to derive ad-hoc embeddings using the embeddings of both substrings {\em and} superstrings from a vocabulary of often large strings. This approach might be comparable to the encoding mechanism of a BLT with a fixed global entropy threshold of $\theta_g$, if our fixed vocabulary is a complete, prefix-free vocabulary\footnote{A vocabulary is complete if every string is a superstring or substring of at least one in-vocab string. A complete, prefix-free vocabulary can be obtained in linear time with a suffix tree.} $V$ of roughly equiprobable strings where e.g.\ $\theta_g \approx \max_{v \in V} -p(v) \log p(v)$.

\newpage

\section*{Limitations}

We use the cosine similarity between the averaged last hidden states of the compressed and uncompressed inputs to measure performance. This has several problems:
\begin{enumerate}
    \item A cosine similarity of 0.999 in isolation is meaningless. If the LM maps every sentence to a single embedding then everything will have a cosine similarity of 1.0. A better metric would likely be the rank of the compressed sentence in the list of the original sentence's nearest neighboring sentences from some large corpus, or the average of each one's rank in the list of the other's.
    \item Cosine similarity is not necessarily aligned with performance on real downstream tasks, although we hope that it is correlated.
    \item Averaging the last hidden states is a fairly aggressive operation. An alternative would be to evaluate the impact of compression using an encoder-only model from the start.
\end{enumerate}

There is also a "vote-splitting" problem at the output end that could hurt performance on downstream tasks. Augmenting the vocabulary, rather than replacing it, means that the model will end up splitting probability mass between the original sub-character tokens and the new combined tokens. This vote-splitting effect could lead to incorrect tokens being selected. Possible options include:
\begin{enumerate}
    \item Only use compression for input, not output. An upside of this is that the model is not expected to conform its output to this ad-hoc post-training code switch, but can hopefully decipher it and then proceed in its usual way.
    \item Remove merged sub-character tokens from the output matrix. The downside is that this would preclude them from being used in appropriate contexts, for example as tokens in other languages.
    \item A better option might be a winner-takes-all approach to vote-splitting, so that a merged token and its sub-character prefix would have all their combined mass given to the one that had more to begin with. This determination could of course be tuned to e.g.\ bias in favor of the merged token, since the model's internal structure likely favors the original sub-character prefix.
\end{enumerate}

\bibpunct{(}{)}{;}{a}{,}{,}
\setcitestyle{authoryear, maxbibnames=2}
\bibliography{custom}

\clearpage

\section{Appendix}

\begin{figure}[H]
    \centering
    \includegraphics[width=\linewidth]{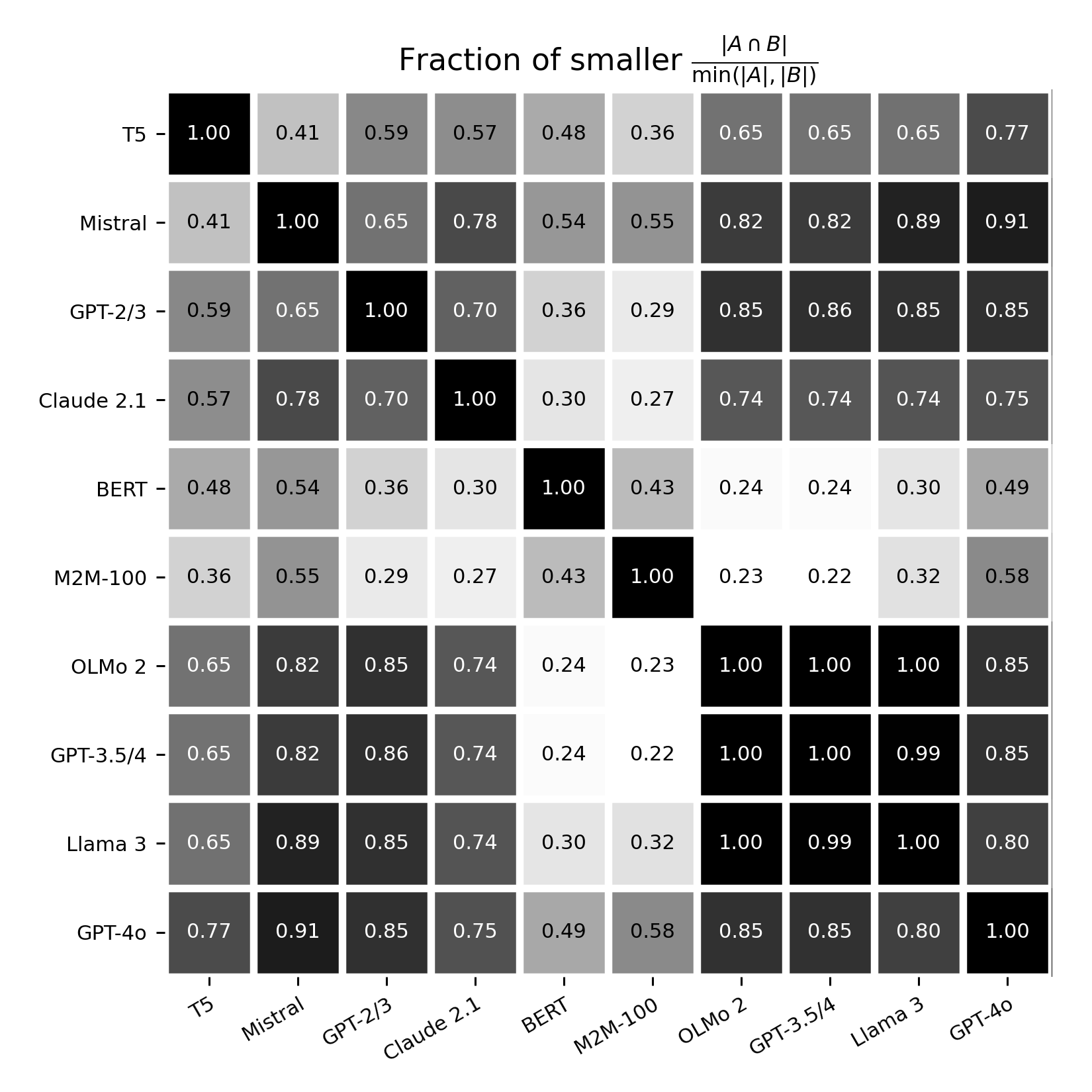}
    \caption{Fraction of smaller normalized vocabulary included in larger normalized vocabulary.}
    \label{fig:tokenizers/PERCENT_SMALLER}
\end{figure}

\end{document}

%% file: res/FLORES-200/tgt_to_src_token_ratios_per_tokenizer_color.tex
\begin{tabular}{lrrrrrrrrrr}
 & Claude 2.1 & OLMo 2 & BERT & Llama 3 & Mistral & T5 & M2M-100 & GPT-2/3 & GPT-3.5/4 & GPT-4o \\
Language &  &  &  &  &  &  &  &  &  &  \\
\texttt{eng (Latn)} & {\cellcolor[HTML]{FFEEE7}} \color[HTML]{000000} 1.00 & {\cellcolor[HTML]{FFEEE7}} \color[HTML]{000000} 1.00 & {\cellcolor[HTML]{FFEEE7}} \color[HTML]{000000} 1.00 & {\cellcolor[HTML]{FFEEE7}} \color[HTML]{000000} 1.00 & {\cellcolor[HTML]{FFEEE7}} \color[HTML]{000000} 1.00 & {\cellcolor[HTML]{FFEEE7}} \color[HTML]{000000} 1.00 & {\cellcolor[HTML]{FFEEE7}} \color[HTML]{000000} 1.00 & {\cellcolor[HTML]{FFEEE7}} \color[HTML]{000000} 1.00 & {\cellcolor[HTML]{FFEEE7}} \color[HTML]{000000} 1.00 & {\cellcolor[HTML]{FFEEE7}} \color[HTML]{000000} 1.00 \\
\texttt{zho (Hans)} & {\cellcolor[HTML]{FEE8DE}} \color[HTML]{000000} 1.70 & {\cellcolor[HTML]{FEE7DB}} \color[HTML]{000000} 1.89 & {\cellcolor[HTML]{FEEAE1}} \color[HTML]{000000} 1.42 & {\cellcolor[HTML]{FFECE3}} \color[HTML]{000000} 1.30 & {\cellcolor[HTML]{FEE8DE}} \color[HTML]{000000} 1.64 & {\cellcolor[HTML]{FFF5F0}} \color[HTML]{000000} 0.29 & {\cellcolor[HTML]{FFEEE6}} \color[HTML]{000000} 1.05 & {\cellcolor[HTML]{FDD7C6}} \color[HTML]{000000} 3.19 & {\cellcolor[HTML]{FEE7DB}} \color[HTML]{000000} 1.89 & {\cellcolor[HTML]{FFECE4}} \color[HTML]{000000} 1.26 \\
\texttt{zho (Hant)} & {\cellcolor[HTML]{FEE6DA}} \color[HTML]{000000} 1.95 & {\cellcolor[HTML]{FEE4D8}} \color[HTML]{000000} 2.16 & {\cellcolor[HTML]{FFEBE2}} \color[HTML]{000000} 1.36 & {\cellcolor[HTML]{FFEBE2}} \color[HTML]{000000} 1.36 & {\cellcolor[HTML]{FEE8DD}} \color[HTML]{000000} 1.77 & {\cellcolor[HTML]{FFF5F0}} \color[HTML]{000000} 0.24 & {\cellcolor[HTML]{FFEEE6}} \color[HTML]{000000} 1.07 & {\cellcolor[HTML]{FED8C7}} \color[HTML]{000000} 3.13 & {\cellcolor[HTML]{FEE4D8}} \color[HTML]{000000} 2.16 & {\cellcolor[HTML]{FEEAE1}} \color[HTML]{000000} 1.42 \\
\texttt{vie (Latn)} & {\cellcolor[HTML]{FDD1BE}} \color[HTML]{000000} 3.56 & {\cellcolor[HTML]{FEE1D4}} \color[HTML]{000000} 2.47 & {\cellcolor[HTML]{FFECE3}} \color[HTML]{000000} 1.30 & {\cellcolor[HTML]{FFEBE2}} \color[HTML]{000000} 1.40 & {\cellcolor[HTML]{FEDACA}} \color[HTML]{000000} 2.98 & {\cellcolor[HTML]{FDCEBB}} \color[HTML]{000000} 3.72 & {\cellcolor[HTML]{FFEDE5}} \color[HTML]{000000} 1.16 & {\cellcolor[HTML]{FCC1A8}} \color[HTML]{000000} 4.60 & {\cellcolor[HTML]{FEE1D4}} \color[HTML]{000000} 2.47 & {\cellcolor[HTML]{FEEAE0}} \color[HTML]{000000} 1.50 \\
\texttt{heb (Hebr)} & {\cellcolor[HTML]{FDD7C6}} \color[HTML]{000000} 3.23 & {\cellcolor[HTML]{FDD0BC}} \color[HTML]{000000} 3.68 & {\cellcolor[HTML]{FEEAE1}} \color[HTML]{000000} 1.43 & {\cellcolor[HTML]{FDD0BC}} \color[HTML]{000000} 3.69 & {\cellcolor[HTML]{FDD3C1}} \color[HTML]{000000} 3.47 & {\cellcolor[HTML]{FFECE3}} \color[HTML]{000000} 1.30 & {\cellcolor[HTML]{FFECE4}} \color[HTML]{000000} 1.23 & {\cellcolor[HTML]{FCC4AD}} \color[HTML]{000000} 4.39 & {\cellcolor[HTML]{FDD0BC}} \color[HTML]{000000} 3.68 & {\cellcolor[HTML]{FEEAE0}} \color[HTML]{000000} 1.49 \\
\texttt{urd (Arab)} & {\cellcolor[HTML]{FCB398}} \color[HTML]{000000} 5.41 & {\cellcolor[HTML]{FCC4AD}} \color[HTML]{000000} 4.43 & {\cellcolor[HTML]{FEE8DD}} \color[HTML]{000000} 1.72 & {\cellcolor[HTML]{FED8C7}} \color[HTML]{000000} 3.13 & {\cellcolor[HTML]{FCC4AD}} \color[HTML]{000000} 4.40 & {\cellcolor[HTML]{FEE7DB}} \color[HTML]{000000} 1.92 & {\cellcolor[HTML]{FFECE3}} \color[HTML]{000000} 1.31 & {\cellcolor[HTML]{FCA486}} \color[HTML]{000000} 6.33 & {\cellcolor[HTML]{FCC4AD}} \color[HTML]{000000} 4.43 & {\cellcolor[HTML]{FEE8DE}} \color[HTML]{000000} 1.68 \\
\texttt{hin (Deva)} & {\cellcolor[HTML]{FCB89E}} \color[HTML]{000000} 5.16 & {\cellcolor[HTML]{FCBDA4}} \color[HTML]{000000} 4.83 & {\cellcolor[HTML]{FEE8DD}} \color[HTML]{000000} 1.78 & {\cellcolor[HTML]{FEE1D3}} \color[HTML]{000000} 2.56 & {\cellcolor[HTML]{FCC1A8}} \color[HTML]{000000} 4.65 & {\cellcolor[HTML]{FEE7DC}} \color[HTML]{000000} 1.84 & {\cellcolor[HTML]{FFEBE2}} \color[HTML]{000000} 1.37 & {\cellcolor[HTML]{FC8F6F}} \color[HTML]{000000} 7.51 & {\cellcolor[HTML]{FCBDA4}} \color[HTML]{000000} 4.83 & {\cellcolor[HTML]{FEE9DF}} \color[HTML]{000000} 1.59 \\
\texttt{ydd (Hebr)} & {\cellcolor[HTML]{FCB79C}} \color[HTML]{000000} 5.20 & {\cellcolor[HTML]{FCB095}} \color[HTML]{000000} 5.59 & {\cellcolor[HTML]{FEDFD0}} \color[HTML]{000000} 2.69 & {\cellcolor[HTML]{FCB095}} \color[HTML]{000000} 5.59 & {\cellcolor[HTML]{FCBDA4}} \color[HTML]{000000} 4.82 & {\cellcolor[HTML]{FEE8DE}} \color[HTML]{000000} 1.69 & {\cellcolor[HTML]{FEE9DF}} \color[HTML]{000000} 1.62 & {\cellcolor[HTML]{FC9D7F}} \color[HTML]{000000} 6.65 & {\cellcolor[HTML]{FCB095}} \color[HTML]{000000} 5.59 & {\cellcolor[HTML]{FEE5D9}} \color[HTML]{000000} 2.04 \\
\texttt{ben (Beng)} & {\cellcolor[HTML]{FC7F5F}} \color[HTML]{F1F1F1} 8.43 & {\cellcolor[HTML]{FCAB8F}} \color[HTML]{000000} 5.89 & {\cellcolor[HTML]{FEE7DB}} \color[HTML]{000000} 1.90 & {\cellcolor[HTML]{FCAB8F}} \color[HTML]{000000} 5.85 & {\cellcolor[HTML]{FCBBA1}} \color[HTML]{000000} 5.01 & {\cellcolor[HTML]{FFEBE2}} \color[HTML]{000000} 1.39 & {\cellcolor[HTML]{FEE9DF}} \color[HTML]{000000} 1.57 & {\cellcolor[HTML]{FB694A}} \color[HTML]{F1F1F1} 9.70 & {\cellcolor[HTML]{FCAB8F}} \color[HTML]{000000} 5.89 & {\cellcolor[HTML]{FEE8DD}} \color[HTML]{000000} 1.72 \\
\texttt{amh (Ethi)} & {\cellcolor[HTML]{FC8262}} \color[HTML]{F1F1F1} 8.25 & {\cellcolor[HTML]{FC8A6A}} \color[HTML]{F1F1F1} 7.77 & {\cellcolor[HTML]{FFF1EA}} \color[HTML]{000000} 0.71 & {\cellcolor[HTML]{FC8A6A}} \color[HTML]{F1F1F1} 7.77 & {\cellcolor[HTML]{FC9E80}} \color[HTML]{000000} 6.64 & {\cellcolor[HTML]{FFECE4}} \color[HTML]{000000} 1.22 & {\cellcolor[HTML]{FEEAE1}} \color[HTML]{000000} 1.43 & {\cellcolor[HTML]{FC8969}} \color[HTML]{F1F1F1} 7.87 & {\cellcolor[HTML]{FC8A6A}} \color[HTML]{F1F1F1} 7.77 & {\cellcolor[HTML]{FCAB8F}} \color[HTML]{000000} 5.91 \\
\texttt{hye (Armn)} & {\cellcolor[HTML]{FA6547}} \color[HTML]{F1F1F1} 9.94 & {\cellcolor[HTML]{F96346}} \color[HTML]{F1F1F1} 10.02 & {\cellcolor[HTML]{FEE5D8}} \color[HTML]{000000} 2.10 & {\cellcolor[HTML]{FB7A5A}} \color[HTML]{F1F1F1} 8.77 & {\cellcolor[HTML]{FCB398}} \color[HTML]{000000} 5.41 & {\cellcolor[HTML]{FEEAE1}} \color[HTML]{000000} 1.42 & {\cellcolor[HTML]{FEEAE0}} \color[HTML]{000000} 1.50 & {\cellcolor[HTML]{F96245}} \color[HTML]{F1F1F1} 10.04 & {\cellcolor[HTML]{F96346}} \color[HTML]{F1F1F1} 10.02 & {\cellcolor[HTML]{FEE7DC}} \color[HTML]{000000} 1.80 \\
\texttt{tam (Taml)} & {\cellcolor[HTML]{F7593F}} \color[HTML]{F1F1F1} 10.52 & {\cellcolor[HTML]{FC8B6B}} \color[HTML]{F1F1F1} 7.74 & {\cellcolor[HTML]{FEE5D8}} \color[HTML]{000000} 2.13 & {\cellcolor[HTML]{FC8B6B}} \color[HTML]{F1F1F1} 7.75 & {\cellcolor[HTML]{FCAA8D}} \color[HTML]{000000} 5.96 & {\cellcolor[HTML]{FFECE4}} \color[HTML]{000000} 1.25 & {\cellcolor[HTML]{FEEAE0}} \color[HTML]{000000} 1.56 & {\cellcolor[HTML]{B51318}} \color[HTML]{F1F1F1} 15.71 & {\cellcolor[HTML]{FC8B6B}} \color[HTML]{F1F1F1} 7.74 & {\cellcolor[HTML]{FEE6DA}} \color[HTML]{000000} 2.00 \\
\texttt{tel (Telu)} & {\cellcolor[HTML]{FA6648}} \color[HTML]{F1F1F1} 9.83 & {\cellcolor[HTML]{FC7F5F}} \color[HTML]{F1F1F1} 8.44 & {\cellcolor[HTML]{FEE4D8}} \color[HTML]{000000} 2.15 & {\cellcolor[HTML]{FC7F5F}} \color[HTML]{F1F1F1} 8.44 & {\cellcolor[HTML]{FC9272}} \color[HTML]{000000} 7.38 & {\cellcolor[HTML]{FFECE3}} \color[HTML]{000000} 1.28 & {\cellcolor[HTML]{FEEAE1}} \color[HTML]{000000} 1.47 & {\cellcolor[HTML]{DD2A25}} \color[HTML]{F1F1F1} 13.18 & {\cellcolor[HTML]{FC7F5F}} \color[HTML]{F1F1F1} 8.44 & {\cellcolor[HTML]{FEE6DA}} \color[HTML]{000000} 1.96 \\
\texttt{shn (Mymr)} & {\cellcolor[HTML]{F75C41}} \color[HTML]{F1F1F1} 10.36 & {\cellcolor[HTML]{BC141A}} \color[HTML]{F1F1F1} 15.33 & {\cellcolor[HTML]{FFF3ED}} \color[HTML]{000000} 0.49 & {\cellcolor[HTML]{BE151A}} \color[HTML]{F1F1F1} 15.13 & {\cellcolor[HTML]{E53228}} \color[HTML]{F1F1F1} 12.65 & {\cellcolor[HTML]{FFF1EA}} \color[HTML]{000000} 0.72 & {\cellcolor[HTML]{FCBFA7}} \color[HTML]{000000} 4.72 & {\cellcolor[HTML]{67000D}} \color[HTML]{F1F1F1} 19.09 & {\cellcolor[HTML]{BC141A}} \color[HTML]{F1F1F1} 15.33 & {\cellcolor[HTML]{FC8565}} \color[HTML]{F1F1F1} 8.10 \\
\end{tabular}

%% file: res/tokenizers/tokenizer_vocab_breakdown.tex
\begin{tabular}{lrr|rrrr|rrrrrrrr}
\toprule
Tokenizer & \makecell{Clean\\vocab\\size} & \makecell{Distinct\\Unicode\\blocks} & \makecell{1 byte\\chars} & \makecell{2 byte\\chars} & \makecell{3 byte\\chars} & \makecell{4 byte\\chars} & \makecell{1 byte\\toks} & \makecell{2 byte\\toks} & \makecell{3 byte\\toks} & \makecell{4 byte\\toks} & \makecell{5 byte\\toks} & \makecell{6 byte\\toks} & \makecell{7 byte\\toks} & \makecell{>7 byte\\toks} \\
          &                               & $\leq$338                                 & $\leq$128 & $\leq$1,920 & $\leq$61,440 & $\leq$1,048,576 \\
\midrule
T5 & 32100 & 15 & 89 & 44 & 10 & 0 & 88 & 812 & 2482 & 3315 & 3914 & 4329 & 4321 & 12839 \\
Mistral & 32515 & 80 & 128 & 580 & 2531 & 59 & 128 & 2072 & 6362 & 5386 & 4658 & 3953 & 2999 & 6957 \\
GPT-2/3 & 50256 & 41 & 128 & 132 & 195 & 1 & 256 & 1916 & 5212 & 7221 & 7305 & 6456 & 5912 & 15978 \\
Claude 2.1 & 64298 & 49 & 128 & 224 & 1079 & 4 & 128 & 2702 & 9391 & 10417 & 8836 & 8076 & 6741 & 18007 \\
OLMo 2 & 99579 & 57 & 128 & 268 & 1056 & 3 & 128 & 3412 & 11786 & 15020 & 14773 & 13144 & 11506 & 29810 \\
GPT-3.5/4 & 100256 & 57 & 128 & 268 & 1056 & 3 & 256 & 3830 & 11939 & 15057 & 14792 & 13100 & 11504 & 29778 \\
BERT & 119543 & 76 & 94 & 649 & 9254 & 2 & 93 & 1458 & 14397 & 21058 & 11266 & 15375 & 11234 & 44662 \\
Llama 3 & 127040 & 68 & 128 & 443 & 3639 & 3 & 128 & 3614 & 14945 & 18497 & 17204 & 19428 & 15068 & 38156 \\
M2M-100 & 128104 & 78 & 96 & 791 & 8244 & 125 & 95 & 2239 & 14637 & 17508 & 16966 & 21918 & 14905 & 39835 \\
GPT-4o & 199998 & 83 & 128 & 657 & 4561 & 23 & 256 & 4653 & 18185 & 26166 & 26796 & 31386 & 24918 & 67638 \\
\bottomrule
\end{tabular}

%% file: res/llm_evaluation/compare_doc_embeddings_color.tex
\begin{tabular}{rrrr|rrrrrrrrrrrr}
& & & & \multicolumn{12}{c}{Language plus \% of new tokens in encodings} \\
\makecell{Layer} & Strategy & $k$ & Mean & \makecell{amh\\79\%} & \makecell{ben\\84\%} & \makecell{heb\\79\%} & \makecell{hin\\78\%} & \makecell{hye\\84\%} & \makecell{shn\\93\%} & \makecell{tam\\87\%} & \makecell{tel\\84\%} & \makecell{urd\\78\%} & \makecell{vie\\26\%} & \makecell{ydd\\81\%} & \makecell{zho\\93\%} \\
\hline

\multirow[c]{10}{*}{0} & \multirow[c]{5}{*}{KNN} & 1 & {\cellcolor[HTML]{FCA588}} \color[HTML]{000000} .644 & {\cellcolor[HTML]{A60F15}} \color[HTML]{F1F1F1} .059 & {\cellcolor[HTML]{FB7C5C}} \color[HTML]{F1F1F1} .513 & {\cellcolor[HTML]{FEE5D9}} \color[HTML]{000000} .884 & {\cellcolor[HTML]{FEE3D6}} \color[HTML]{000000} .868 & {\cellcolor[HTML]{FCB095}} \color[HTML]{000000} .681 & {\cellcolor[HTML]{CC191E}} \color[HTML]{F1F1F1} .193 & {\cellcolor[HTML]{FB694A}} \color[HTML]{F1F1F1} .451 & {\cellcolor[HTML]{FB7252}} \color[HTML]{F1F1F1} .480 & {\cellcolor[HTML]{FEE1D3}} \color[HTML]{000000} .853 & {\cellcolor[HTML]{FEE0D2}} \color[HTML]{000000} .848 & {\cellcolor[HTML]{FFEBE2}} \color[HTML]{000000} .922 & {\cellcolor[HTML]{FFF4EF}} \color[HTML]{000000} .977 \\
 &  & 2 & {\cellcolor[HTML]{FCBCA2}} \color[HTML]{000000} .719 & {\cellcolor[HTML]{A30F15}} \color[HTML]{F1F1F1} .051 & {\cellcolor[HTML]{FCBDA4}} \color[HTML]{000000} .723 & {\cellcolor[HTML]{FEE3D7}} \color[HTML]{000000} .870 & {\cellcolor[HTML]{FEE3D6}} \color[HTML]{000000} .868 & {\cellcolor[HTML]{FDD1BE}} \color[HTML]{000000} .796 & {\cellcolor[HTML]{DB2824}} \color[HTML]{F1F1F1} .246 & {\cellcolor[HTML]{FCBFA7}} \color[HTML]{000000} .734 & {\cellcolor[HTML]{FDCBB6}} \color[HTML]{000000} .774 & {\cellcolor[HTML]{FDD7C6}} \color[HTML]{000000} .815 & {\cellcolor[HTML]{FEE0D2}} \color[HTML]{000000} .849 & {\cellcolor[HTML]{FEEAE1}} \color[HTML]{000000} .916 & {\cellcolor[HTML]{FFF5F0}} \color[HTML]{000000} .982 \\
 &  & 3 & {\cellcolor[HTML]{FCAA8D}} \color[HTML]{000000} .663 & {\cellcolor[HTML]{9D0D14}} \color[HTML]{F1F1F1} .036 & {\cellcolor[HTML]{FC8F6F}} \color[HTML]{000000} .573 & {\cellcolor[HTML]{FEE1D4}} \color[HTML]{000000} .859 & {\cellcolor[HTML]{FEE3D6}} \color[HTML]{000000} .868 & {\cellcolor[HTML]{FCAB8F}} \color[HTML]{000000} .666 & {\cellcolor[HTML]{D01D1F}} \color[HTML]{F1F1F1} .203 & {\cellcolor[HTML]{FC9C7D}} \color[HTML]{000000} .618 & {\cellcolor[HTML]{FCA285}} \color[HTML]{000000} .639 & {\cellcolor[HTML]{FDCCB8}} \color[HTML]{000000} .780 & {\cellcolor[HTML]{FEE0D2}} \color[HTML]{000000} .848 & {\cellcolor[HTML]{FEE7DB}} \color[HTML]{000000} .890 & {\cellcolor[HTML]{FFF4EF}} \color[HTML]{000000} .978 \\
 &  & 5 & {\cellcolor[HTML]{FC7F5F}} \color[HTML]{F1F1F1} .522 & {\cellcolor[HTML]{900A12}} \color[HTML]{F1F1F1} .009 & {\cellcolor[HTML]{F14130}} \color[HTML]{F1F1F1} .338 & {\cellcolor[HTML]{FDD5C4}} \color[HTML]{000000} .812 & {\cellcolor[HTML]{FEE3D6}} \color[HTML]{000000} .868 & {\cellcolor[HTML]{EF3C2C}} \color[HTML]{F1F1F1} .320 & {\cellcolor[HTML]{9C0D14}} \color[HTML]{F1F1F1} .032 & {\cellcolor[HTML]{DC2924}} \color[HTML]{F1F1F1} .251 & {\cellcolor[HTML]{D92523}} \color[HTML]{F1F1F1} .236 & {\cellcolor[HTML]{FDC6B0}} \color[HTML]{000000} .759 & {\cellcolor[HTML]{FEE0D2}} \color[HTML]{000000} .848 & {\cellcolor[HTML]{FDD7C6}} \color[HTML]{000000} .816 & {\cellcolor[HTML]{FFF3ED}} \color[HTML]{000000} .972 \\
 &  & 10 & {\cellcolor[HTML]{F5533B}} \color[HTML]{F1F1F1} .385 & {\cellcolor[HTML]{8E0912}} \color[HTML]{F1F1F1} .002 & {\cellcolor[HTML]{B11218}} \color[HTML]{F1F1F1} .096 & {\cellcolor[HTML]{FCB095}} \color[HTML]{000000} .684 & {\cellcolor[HTML]{FEE3D6}} \color[HTML]{000000} .868 & {\cellcolor[HTML]{8C0912}} \color[HTML]{F1F1F1} .001 & {\cellcolor[HTML]{6D010E}} \color[HTML]{F1F1F1} -.069 & {\cellcolor[HTML]{8E0912}} \color[HTML]{F1F1F1} .005 & {\cellcolor[HTML]{67000D}} \color[HTML]{F1F1F1} -.082 & {\cellcolor[HTML]{FCBCA2}} \color[HTML]{000000} .722 & {\cellcolor[HTML]{FEE0D2}} \color[HTML]{000000} .848 & {\cellcolor[HTML]{FC8F6F}} \color[HTML]{000000} .574 & {\cellcolor[HTML]{FFF3ED}} \color[HTML]{000000} .970 \\
 & LinReg &  & {\cellcolor[HTML]{FDD4C2}} \color[HTML]{000000} .809 & {\cellcolor[HTML]{FC9E80}} \color[HTML]{000000} .624 & {\cellcolor[HTML]{FCBDA4}} \color[HTML]{000000} .723 & {\cellcolor[HTML]{FEE3D7}} \color[HTML]{000000} .870 & {\cellcolor[HTML]{FEE3D6}} \color[HTML]{000000} .868 & {\cellcolor[HTML]{FDD1BE}} \color[HTML]{000000} .796 & {\cellcolor[HTML]{FDC7B2}} \color[HTML]{000000} .762 & {\cellcolor[HTML]{FCBFA7}} \color[HTML]{000000} .734 & {\cellcolor[HTML]{FDCBB6}} \color[HTML]{000000} .774 & {\cellcolor[HTML]{FDD7C6}} \color[HTML]{000000} .815 & {\cellcolor[HTML]{FEE0D2}} \color[HTML]{000000} .849 & {\cellcolor[HTML]{FEEAE1}} \color[HTML]{000000} .916 & {\cellcolor[HTML]{FFF5F0}} \color[HTML]{000000} .982 \\
 & \multirow[c]{4}{*}{LocalLinReg} & 2 & {\cellcolor[HTML]{FCBCA2}} \color[HTML]{000000} .719 & {\cellcolor[HTML]{A50F15}} \color[HTML]{F1F1F1} .052 & {\cellcolor[HTML]{FCBDA4}} \color[HTML]{000000} .723 & {\cellcolor[HTML]{FEE3D7}} \color[HTML]{000000} .870 & {\cellcolor[HTML]{FEE3D6}} \color[HTML]{000000} .868 & {\cellcolor[HTML]{FDD1BE}} \color[HTML]{000000} .796 & {\cellcolor[HTML]{DB2824}} \color[HTML]{F1F1F1} .246 & {\cellcolor[HTML]{FCBFA7}} \color[HTML]{000000} .734 & {\cellcolor[HTML]{FDCBB6}} \color[HTML]{000000} .774 & {\cellcolor[HTML]{FDD7C6}} \color[HTML]{000000} .815 & {\cellcolor[HTML]{FEE0D2}} \color[HTML]{000000} .849 & {\cellcolor[HTML]{FEEAE1}} \color[HTML]{000000} .916 & {\cellcolor[HTML]{FFF5F0}} \color[HTML]{000000} .982 \\
 &  & 3 & {\cellcolor[HTML]{FCBCA2}} \color[HTML]{000000} .719 & {\cellcolor[HTML]{A50F15}} \color[HTML]{F1F1F1} .052 & {\cellcolor[HTML]{FCBDA4}} \color[HTML]{000000} .723 & {\cellcolor[HTML]{FEE3D7}} \color[HTML]{000000} .870 & {\cellcolor[HTML]{FEE3D6}} \color[HTML]{000000} .868 & {\cellcolor[HTML]{FDD1BE}} \color[HTML]{000000} .796 & {\cellcolor[HTML]{DB2824}} \color[HTML]{F1F1F1} .247 & {\cellcolor[HTML]{FCBFA7}} \color[HTML]{000000} .734 & {\cellcolor[HTML]{FDCBB6}} \color[HTML]{000000} .774 & {\cellcolor[HTML]{FDD7C6}} \color[HTML]{000000} .815 & {\cellcolor[HTML]{FEE0D2}} \color[HTML]{000000} .849 & {\cellcolor[HTML]{FEEAE1}} \color[HTML]{000000} .916 & {\cellcolor[HTML]{FFF5F0}} \color[HTML]{000000} .982 \\
 &  & 5 & {\cellcolor[HTML]{FCBCA2}} \color[HTML]{000000} .719 & {\cellcolor[HTML]{A50F15}} \color[HTML]{F1F1F1} .052 & {\cellcolor[HTML]{FCBDA4}} \color[HTML]{000000} .723 & {\cellcolor[HTML]{FEE3D7}} \color[HTML]{000000} .870 & {\cellcolor[HTML]{FEE3D6}} \color[HTML]{000000} .868 & {\cellcolor[HTML]{FDD1BE}} \color[HTML]{000000} .796 & {\cellcolor[HTML]{DC2924}} \color[HTML]{F1F1F1} .250 & {\cellcolor[HTML]{FCBFA7}} \color[HTML]{000000} .734 & {\cellcolor[HTML]{FDCBB6}} \color[HTML]{000000} .774 & {\cellcolor[HTML]{FDD7C6}} \color[HTML]{000000} .815 & {\cellcolor[HTML]{FEE0D2}} \color[HTML]{000000} .849 & {\cellcolor[HTML]{FEEAE1}} \color[HTML]{000000} .916 & {\cellcolor[HTML]{FFF5F0}} \color[HTML]{000000} .982 \\
 &  & 10 & {\cellcolor[HTML]{FCBCA2}} \color[HTML]{000000} .719 & {\cellcolor[HTML]{A30F15}} \color[HTML]{F1F1F1} .049 & {\cellcolor[HTML]{FCBDA4}} \color[HTML]{000000} .723 & {\cellcolor[HTML]{FEE3D7}} \color[HTML]{000000} .870 & {\cellcolor[HTML]{FEE3D6}} \color[HTML]{000000} .868 & {\cellcolor[HTML]{FDD1BE}} \color[HTML]{000000} .796 & {\cellcolor[HTML]{DD2A25}} \color[HTML]{F1F1F1} .254 & {\cellcolor[HTML]{FCBFA7}} \color[HTML]{000000} .734 & {\cellcolor[HTML]{FDCBB6}} \color[HTML]{000000} .774 & {\cellcolor[HTML]{FDD7C6}} \color[HTML]{000000} .815 & {\cellcolor[HTML]{FEE0D2}} \color[HTML]{000000} .849 & {\cellcolor[HTML]{FEEAE1}} \color[HTML]{000000} .916 & {\cellcolor[HTML]{FFF5F0}} \color[HTML]{000000} .982 \\
 \hline
\multirow[c]{10}{*}{1} & \multirow[c]{5}{*}{KNN} & 1 & {\cellcolor[HTML]{FCA689}} \color[HTML]{000000} .648 & {\cellcolor[HTML]{DB2824}} \color[HTML]{F1F1F1} .244 & {\cellcolor[HTML]{FC8B6B}} \color[HTML]{F1F1F1} .560 & {\cellcolor[HTML]{FDCCB8}} \color[HTML]{000000} .780 & {\cellcolor[HTML]{FEE3D6}} \color[HTML]{000000} .867 & {\cellcolor[HTML]{FCB499}} \color[HTML]{000000} .694 & {\cellcolor[HTML]{DE2B25}} \color[HTML]{F1F1F1} .258 & {\cellcolor[HTML]{F85F43}} \color[HTML]{F1F1F1} .421 & {\cellcolor[HTML]{FC8E6E}} \color[HTML]{000000} .570 & {\cellcolor[HTML]{FDCEBB}} \color[HTML]{000000} .786 & {\cellcolor[HTML]{FEE0D2}} \color[HTML]{000000} .849 & {\cellcolor[HTML]{FDC7B2}} \color[HTML]{000000} .763 & {\cellcolor[HTML]{FFF5F0}} \color[HTML]{000000} .985 \\
 &  & 2 & {\cellcolor[HTML]{FC9B7C}} \color[HTML]{000000} .614 & {\cellcolor[HTML]{CF1C1F}} \color[HTML]{F1F1F1} .201 & {\cellcolor[HTML]{FB7C5C}} \color[HTML]{F1F1F1} .510 & {\cellcolor[HTML]{FCC1A8}} \color[HTML]{000000} .736 & {\cellcolor[HTML]{FEE3D6}} \color[HTML]{000000} .867 & {\cellcolor[HTML]{FC8767}} \color[HTML]{F1F1F1} .548 & {\cellcolor[HTML]{D72322}} \color[HTML]{F1F1F1} .228 & {\cellcolor[HTML]{F75B40}} \color[HTML]{F1F1F1} .407 & {\cellcolor[HTML]{FC8060}} \color[HTML]{F1F1F1} .523 & {\cellcolor[HTML]{FDD0BC}} \color[HTML]{000000} .792 & {\cellcolor[HTML]{FEE0D2}} \color[HTML]{000000} .849 & {\cellcolor[HTML]{FCBCA2}} \color[HTML]{000000} .721 & {\cellcolor[HTML]{FFF5F0}} \color[HTML]{000000} .982 \\
 &  & 3 & {\cellcolor[HTML]{FC9474}} \color[HTML]{000000} .592 & {\cellcolor[HTML]{E43027}} \color[HTML]{F1F1F1} .279 & {\cellcolor[HTML]{FB7757}} \color[HTML]{F1F1F1} .496 & {\cellcolor[HTML]{FCB296}} \color[HTML]{000000} .687 & {\cellcolor[HTML]{FEE3D6}} \color[HTML]{000000} .867 & {\cellcolor[HTML]{F85D42}} \color[HTML]{F1F1F1} .415 & {\cellcolor[HTML]{E43027}} \color[HTML]{F1F1F1} .277 & {\cellcolor[HTML]{F34A36}} \color[HTML]{F1F1F1} .363 & {\cellcolor[HTML]{FA6849}} \color[HTML]{F1F1F1} .447 & {\cellcolor[HTML]{FDCEBB}} \color[HTML]{000000} .785 & {\cellcolor[HTML]{FEE0D2}} \color[HTML]{000000} .848 & {\cellcolor[HTML]{FCA98C}} \color[HTML]{000000} .658 & {\cellcolor[HTML]{FFF4EF}} \color[HTML]{000000} .978 \\
 &  & 5 & {\cellcolor[HTML]{FC8F6F}} \color[HTML]{000000} .575 & {\cellcolor[HTML]{EE3A2C}} \color[HTML]{F1F1F1} .315 & {\cellcolor[HTML]{FB694A}} \color[HTML]{F1F1F1} .449 & {\cellcolor[HTML]{FCBFA7}} \color[HTML]{000000} .733 & {\cellcolor[HTML]{FEE3D6}} \color[HTML]{000000} .867 & {\cellcolor[HTML]{F24633}} \color[HTML]{F1F1F1} .349 & {\cellcolor[HTML]{D82422}} \color[HTML]{F1F1F1} .232 & {\cellcolor[HTML]{E12D26}} \color[HTML]{F1F1F1} .264 & {\cellcolor[HTML]{F4503A}} \color[HTML]{F1F1F1} .379 & {\cellcolor[HTML]{FDCBB6}} \color[HTML]{000000} .773 & {\cellcolor[HTML]{FEE0D2}} \color[HTML]{000000} .848 & {\cellcolor[HTML]{FCB99F}} \color[HTML]{000000} .710 & {\cellcolor[HTML]{FFF4EE}} \color[HTML]{000000} .974 \\
 &  & 10 & {\cellcolor[HTML]{FC8767}} \color[HTML]{F1F1F1} .549 & {\cellcolor[HTML]{E02C26}} \color[HTML]{F1F1F1} .260 & {\cellcolor[HTML]{F5533B}} \color[HTML]{F1F1F1} .385 & {\cellcolor[HTML]{FCBDA4}} \color[HTML]{000000} .725 & {\cellcolor[HTML]{FEE3D6}} \color[HTML]{000000} .868 & {\cellcolor[HTML]{E83429}} \color[HTML]{F1F1F1} .291 & {\cellcolor[HTML]{DD2A25}} \color[HTML]{F1F1F1} .252 & {\cellcolor[HTML]{D21F20}} \color[HTML]{F1F1F1} .212 & {\cellcolor[HTML]{E43027}} \color[HTML]{F1F1F1} .278 & {\cellcolor[HTML]{FDC9B3}} \color[HTML]{000000} .766 & {\cellcolor[HTML]{FEE0D2}} \color[HTML]{000000} .848 & {\cellcolor[HTML]{FCC1A8}} \color[HTML]{000000} .738 & {\cellcolor[HTML]{FFF3ED}} \color[HTML]{000000} .969 \\
 & LinReg &  & {\cellcolor[HTML]{FCBDA4}} \color[HTML]{000000} .724 & {\cellcolor[HTML]{FDCBB6}} \color[HTML]{000000} .773 & {\cellcolor[HTML]{FC8E6E}} \color[HTML]{000000} .572 & {\cellcolor[HTML]{FCBFA7}} \color[HTML]{000000} .731 & {\cellcolor[HTML]{FDD7C6}} \color[HTML]{000000} .818 & {\cellcolor[HTML]{FCA183}} \color[HTML]{000000} .631 & {\cellcolor[HTML]{FDC5AE}} \color[HTML]{000000} .755 & {\cellcolor[HTML]{FC9B7C}} \color[HTML]{000000} .612 & {\cellcolor[HTML]{FCB499}} \color[HTML]{000000} .695 & {\cellcolor[HTML]{FCA588}} \color[HTML]{000000} .647 & {\cellcolor[HTML]{FDD4C2}} \color[HTML]{000000} .809 & {\cellcolor[HTML]{FDC7B2}} \color[HTML]{000000} .762 & {\cellcolor[HTML]{FEE6DA}} \color[HTML]{000000} .887 \\
 & \multirow[c]{4}{*}{LocalLinReg} & 2 & {\cellcolor[HTML]{FC9C7D}} \color[HTML]{000000} .615 & {\cellcolor[HTML]{CF1C1F}} \color[HTML]{F1F1F1} .201 & {\cellcolor[HTML]{FB7C5C}} \color[HTML]{F1F1F1} .514 & {\cellcolor[HTML]{FCC1A8}} \color[HTML]{000000} .736 & {\cellcolor[HTML]{FEE3D6}} \color[HTML]{000000} .867 & {\cellcolor[HTML]{FC8A6A}} \color[HTML]{F1F1F1} .556 & {\cellcolor[HTML]{D72322}} \color[HTML]{F1F1F1} .229 & {\cellcolor[HTML]{F75C41}} \color[HTML]{F1F1F1} .410 & {\cellcolor[HTML]{FC8060}} \color[HTML]{F1F1F1} .523 & {\cellcolor[HTML]{FDD0BC}} \color[HTML]{000000} .792 & {\cellcolor[HTML]{FEE0D2}} \color[HTML]{000000} .849 & {\cellcolor[HTML]{FCBCA2}} \color[HTML]{000000} .721 & {\cellcolor[HTML]{FFF5F0}} \color[HTML]{000000} .983 \\
 &  & 3 & {\cellcolor[HTML]{FC9777}} \color[HTML]{000000} .601 & {\cellcolor[HTML]{DB2824}} \color[HTML]{F1F1F1} .244 & {\cellcolor[HTML]{FB7B5B}} \color[HTML]{F1F1F1} .509 & {\cellcolor[HTML]{FCB095}} \color[HTML]{000000} .681 & {\cellcolor[HTML]{FEE3D6}} \color[HTML]{000000} .867 & {\cellcolor[HTML]{FC8D6D}} \color[HTML]{F1F1F1} .565 & {\cellcolor[HTML]{DE2B25}} \color[HTML]{F1F1F1} .257 & {\cellcolor[HTML]{F34C37}} \color[HTML]{F1F1F1} .367 & {\cellcolor[HTML]{FB694A}} \color[HTML]{F1F1F1} .449 & {\cellcolor[HTML]{FDD0BC}} \color[HTML]{000000} .791 & {\cellcolor[HTML]{FEE0D2}} \color[HTML]{000000} .848 & {\cellcolor[HTML]{FCA98C}} \color[HTML]{000000} .657 & {\cellcolor[HTML]{FFF5F0}} \color[HTML]{000000} .981 \\
 &  & 5 & {\cellcolor[HTML]{FC9B7C}} \color[HTML]{000000} .613 & {\cellcolor[HTML]{F34935}} \color[HTML]{F1F1F1} .359 & {\cellcolor[HTML]{FB7858}} \color[HTML]{F1F1F1} .501 & {\cellcolor[HTML]{FCC3AB}} \color[HTML]{000000} .744 & {\cellcolor[HTML]{FEE3D6}} \color[HTML]{000000} .867 & {\cellcolor[HTML]{FC9272}} \color[HTML]{000000} .581 & {\cellcolor[HTML]{DD2A25}} \color[HTML]{F1F1F1} .254 & {\cellcolor[HTML]{EC382B}} \color[HTML]{F1F1F1} .308 & {\cellcolor[HTML]{F6583E}} \color[HTML]{F1F1F1} .399 & {\cellcolor[HTML]{FDD0BC}} \color[HTML]{000000} .791 & {\cellcolor[HTML]{FEE0D2}} \color[HTML]{000000} .848 & {\cellcolor[HTML]{FCBDA4}} \color[HTML]{000000} .726 & {\cellcolor[HTML]{FFF4EF}} \color[HTML]{000000} .979 \\
 &  & 10 & {\cellcolor[HTML]{FC9E80}} \color[HTML]{000000} .624 & {\cellcolor[HTML]{FB7757}} \color[HTML]{F1F1F1} .495 & {\cellcolor[HTML]{FB7252}} \color[HTML]{F1F1F1} .478 & {\cellcolor[HTML]{FCC4AD}} \color[HTML]{000000} .750 & {\cellcolor[HTML]{FEE3D6}} \color[HTML]{000000} .868 & {\cellcolor[HTML]{FC8464}} \color[HTML]{F1F1F1} .538 & {\cellcolor[HTML]{EB372A}} \color[HTML]{F1F1F1} .305 & {\cellcolor[HTML]{E63328}} \color[HTML]{F1F1F1} .287 & {\cellcolor[HTML]{F44F39}} \color[HTML]{F1F1F1} .374 & {\cellcolor[HTML]{FDD2BF}} \color[HTML]{000000} .801 & {\cellcolor[HTML]{FEE0D2}} \color[HTML]{000000} .848 & {\cellcolor[HTML]{FDC9B3}} \color[HTML]{000000} .766 & {\cellcolor[HTML]{FFF4EF}} \color[HTML]{000000} .979 \\
 \hline
\multirow[c]{10}{*}{15} & \multirow[c]{5}{*}{KNN} & 1 & {\cellcolor[HTML]{FC9474}} \color[HTML]{000000} .591 & {\cellcolor[HTML]{F03F2E}} \color[HTML]{F1F1F1} .329 & {\cellcolor[HTML]{FA6648}} \color[HTML]{F1F1F1} .443 & {\cellcolor[HTML]{FDC9B3}} \color[HTML]{000000} .767 & {\cellcolor[HTML]{FEE3D6}} \color[HTML]{000000} .867 & {\cellcolor[HTML]{F34C37}} \color[HTML]{F1F1F1} .368 & {\cellcolor[HTML]{EA362A}} \color[HTML]{F1F1F1} .298 & {\cellcolor[HTML]{D92523}} \color[HTML]{F1F1F1} .238 & {\cellcolor[HTML]{FB6B4B}} \color[HTML]{F1F1F1} .452 & {\cellcolor[HTML]{FCC2AA}} \color[HTML]{000000} .741 & {\cellcolor[HTML]{FEE0D2}} \color[HTML]{000000} .848 & {\cellcolor[HTML]{FDCEBB}} \color[HTML]{000000} .786 & {\cellcolor[HTML]{FFF2EB}} \color[HTML]{000000} .960 \\
 &  & 2 & {\cellcolor[HTML]{FC8969}} \color[HTML]{F1F1F1} .554 & {\cellcolor[HTML]{EB372A}} \color[HTML]{F1F1F1} .303 & {\cellcolor[HTML]{F34C37}} \color[HTML]{F1F1F1} .367 & {\cellcolor[HTML]{FDC7B2}} \color[HTML]{000000} .764 & {\cellcolor[HTML]{FEE3D6}} \color[HTML]{000000} .867 & {\cellcolor[HTML]{E63328}} \color[HTML]{F1F1F1} .288 & {\cellcolor[HTML]{D01D1F}} \color[HTML]{F1F1F1} .202 & {\cellcolor[HTML]{CF1C1F}} \color[HTML]{F1F1F1} .198 & {\cellcolor[HTML]{F34C37}} \color[HTML]{F1F1F1} .364 & {\cellcolor[HTML]{FCBCA2}} \color[HTML]{000000} .720 & {\cellcolor[HTML]{FEE0D2}} \color[HTML]{000000} .848 & {\cellcolor[HTML]{FDC7B2}} \color[HTML]{000000} .763 & {\cellcolor[HTML]{FFF2EB}} \color[HTML]{000000} .960 \\
 &  & 3 & {\cellcolor[HTML]{FC8161}} \color[HTML]{F1F1F1} .529 & {\cellcolor[HTML]{DE2B25}} \color[HTML]{F1F1F1} .259 & {\cellcolor[HTML]{EE3A2C}} \color[HTML]{F1F1F1} .317 & {\cellcolor[HTML]{FDC6B0}} \color[HTML]{000000} .759 & {\cellcolor[HTML]{FEE3D6}} \color[HTML]{000000} .867 & {\cellcolor[HTML]{DD2A25}} \color[HTML]{F1F1F1} .255 & {\cellcolor[HTML]{BE151A}} \color[HTML]{F1F1F1} .141 & {\cellcolor[HTML]{C4161C}} \color[HTML]{F1F1F1} .160 & {\cellcolor[HTML]{EF3C2C}} \color[HTML]{F1F1F1} .319 & {\cellcolor[HTML]{FCBCA2}} \color[HTML]{000000} .720 & {\cellcolor[HTML]{FEE0D2}} \color[HTML]{000000} .848 & {\cellcolor[HTML]{FCC3AB}} \color[HTML]{000000} .744 & {\cellcolor[HTML]{FFF2EB}} \color[HTML]{000000} .960 \\
 &  & 5 & {\cellcolor[HTML]{FB7353}} \color[HTML]{F1F1F1} .482 & {\cellcolor[HTML]{D52221}} \color[HTML]{F1F1F1} .226 & {\cellcolor[HTML]{DD2A25}} \color[HTML]{F1F1F1} .254 & {\cellcolor[HTML]{FCB499}} \color[HTML]{000000} .696 & {\cellcolor[HTML]{FEE3D6}} \color[HTML]{000000} .868 & {\cellcolor[HTML]{C2161B}} \color[HTML]{F1F1F1} .154 & {\cellcolor[HTML]{B91419}} \color[HTML]{F1F1F1} .123 & {\cellcolor[HTML]{C3161B}} \color[HTML]{F1F1F1} .159 & {\cellcolor[HTML]{B81419}} \color[HTML]{F1F1F1} .120 & {\cellcolor[HTML]{FCBFA7}} \color[HTML]{000000} .733 & {\cellcolor[HTML]{FEE0D2}} \color[HTML]{000000} .848 & {\cellcolor[HTML]{FCA689}} \color[HTML]{000000} .649 & {\cellcolor[HTML]{FFF1EA}} \color[HTML]{000000} .958 \\
 &  & 10 & {\cellcolor[HTML]{FA6547}} \color[HTML]{F1F1F1} .438 & {\cellcolor[HTML]{BE151A}} \color[HTML]{F1F1F1} .143 & {\cellcolor[HTML]{C1161B}} \color[HTML]{F1F1F1} .148 & {\cellcolor[HTML]{FCB79C}} \color[HTML]{000000} .704 & {\cellcolor[HTML]{FEE3D6}} \color[HTML]{000000} .867 & {\cellcolor[HTML]{B21218}} \color[HTML]{F1F1F1} .100 & {\cellcolor[HTML]{840711}} \color[HTML]{F1F1F1} -.017 & {\cellcolor[HTML]{AD1117}} \color[HTML]{F1F1F1} .084 & {\cellcolor[HTML]{A10E15}} \color[HTML]{F1F1F1} .044 & {\cellcolor[HTML]{FCB99F}} \color[HTML]{000000} .712 & {\cellcolor[HTML]{FEE0D2}} \color[HTML]{000000} .848 & {\cellcolor[HTML]{FCAA8D}} \color[HTML]{000000} .664 & {\cellcolor[HTML]{FFF1EA}} \color[HTML]{000000} .956 \\
 & LinReg &  & {\cellcolor[HTML]{F34A36}} \color[HTML]{F1F1F1} .360 & {\cellcolor[HTML]{F75B40}} \color[HTML]{F1F1F1} .407 & {\cellcolor[HTML]{F14130}} \color[HTML]{F1F1F1} .338 & {\cellcolor[HTML]{E83429}} \color[HTML]{F1F1F1} .289 & {\cellcolor[HTML]{D92523}} \color[HTML]{F1F1F1} .237 & {\cellcolor[HTML]{FB6B4B}} \color[HTML]{F1F1F1} .452 & {\cellcolor[HTML]{FA6849}} \color[HTML]{F1F1F1} .444 & {\cellcolor[HTML]{F4503A}} \color[HTML]{F1F1F1} .379 & {\cellcolor[HTML]{FB6B4B}} \color[HTML]{F1F1F1} .453 & {\cellcolor[HTML]{CC191E}} \color[HTML]{F1F1F1} .189 & {\cellcolor[HTML]{FCB095}} \color[HTML]{000000} .682 & {\cellcolor[HTML]{F4503A}} \color[HTML]{F1F1F1} .378 & {\cellcolor[HTML]{A91016}} \color[HTML]{F1F1F1} .068 \\
 & \multirow[c]{4}{*}{LocalLinReg} & 2 & {\cellcolor[HTML]{FC9070}} \color[HTML]{000000} .578 & {\cellcolor[HTML]{F0402F}} \color[HTML]{F1F1F1} .331 & {\cellcolor[HTML]{F44F39}} \color[HTML]{F1F1F1} .374 & {\cellcolor[HTML]{FDC9B3}} \color[HTML]{000000} .767 & {\cellcolor[HTML]{FEE3D6}} \color[HTML]{000000} .867 & {\cellcolor[HTML]{F14331}} \color[HTML]{F1F1F1} .341 & {\cellcolor[HTML]{E63328}} \color[HTML]{F1F1F1} .286 & {\cellcolor[HTML]{D52221}} \color[HTML]{F1F1F1} .225 & {\cellcolor[HTML]{FA6849}} \color[HTML]{F1F1F1} .445 & {\cellcolor[HTML]{FCBDA4}} \color[HTML]{000000} .725 & {\cellcolor[HTML]{FEE0D2}} \color[HTML]{000000} .848 & {\cellcolor[HTML]{FDCAB5}} \color[HTML]{000000} .772 & {\cellcolor[HTML]{FFF2EB}} \color[HTML]{000000} .960 \\
 &  & 3 & {\cellcolor[HTML]{FC8F6F}} \color[HTML]{000000} .576 & {\cellcolor[HTML]{F5533B}} \color[HTML]{F1F1F1} .387 & {\cellcolor[HTML]{F44F39}} \color[HTML]{F1F1F1} .375 & {\cellcolor[HTML]{FDC7B2}} \color[HTML]{000000} .762 & {\cellcolor[HTML]{FEE3D6}} \color[HTML]{000000} .867 & {\cellcolor[HTML]{ED392B}} \color[HTML]{F1F1F1} .311 & {\cellcolor[HTML]{E93529}} \color[HTML]{F1F1F1} .295 & {\cellcolor[HTML]{C7171C}} \color[HTML]{F1F1F1} .171 & {\cellcolor[HTML]{FB694A}} \color[HTML]{F1F1F1} .448 & {\cellcolor[HTML]{FCBCA2}} \color[HTML]{000000} .722 & {\cellcolor[HTML]{FEE0D2}} \color[HTML]{000000} .848 & {\cellcolor[HTML]{FDC9B3}} \color[HTML]{000000} .765 & {\cellcolor[HTML]{FFF2EB}} \color[HTML]{000000} .962 \\
 &  & 5 & {\cellcolor[HTML]{FC9E80}} \color[HTML]{000000} .623 & {\cellcolor[HTML]{F85D42}} \color[HTML]{F1F1F1} .418 & {\cellcolor[HTML]{FB7757}} \color[HTML]{F1F1F1} .496 & {\cellcolor[HTML]{FDC6B0}} \color[HTML]{000000} .757 & {\cellcolor[HTML]{FEE3D6}} \color[HTML]{000000} .867 & {\cellcolor[HTML]{F44F39}} \color[HTML]{F1F1F1} .375 & {\cellcolor[HTML]{F7593F}} \color[HTML]{F1F1F1} .405 & {\cellcolor[HTML]{F7593F}} \color[HTML]{F1F1F1} .404 & {\cellcolor[HTML]{FA6648}} \color[HTML]{F1F1F1} .439 & {\cellcolor[HTML]{FCC1A8}} \color[HTML]{000000} .738 & {\cellcolor[HTML]{FEE0D2}} \color[HTML]{000000} .848 & {\cellcolor[HTML]{FDC6B0}} \color[HTML]{000000} .759 & {\cellcolor[HTML]{FFF2EC}} \color[HTML]{000000} .968 \\
 &  & 10 & {\cellcolor[HTML]{FC9C7D}} \color[HTML]{000000} .618 & {\cellcolor[HTML]{EE3A2C}} \color[HTML]{F1F1F1} .318 & {\cellcolor[HTML]{FB6E4E}} \color[HTML]{F1F1F1} .467 & {\cellcolor[HTML]{FDD1BE}} \color[HTML]{000000} .796 & {\cellcolor[HTML]{FEE3D6}} \color[HTML]{000000} .868 & {\cellcolor[HTML]{F6553C}} \color[HTML]{F1F1F1} .392 & {\cellcolor[HTML]{F44D38}} \color[HTML]{F1F1F1} .370 & {\cellcolor[HTML]{F44D38}} \color[HTML]{F1F1F1} .371 & {\cellcolor[HTML]{F96245}} \color[HTML]{F1F1F1} .430 & {\cellcolor[HTML]{FDCDB9}} \color[HTML]{000000} .784 & {\cellcolor[HTML]{FEE0D2}} \color[HTML]{000000} .848 & {\cellcolor[HTML]{FDD2BF}} \color[HTML]{000000} .800 & {\cellcolor[HTML]{FFF3ED}} \color[HTML]{000000} .969 \\
 \hline
\multirow[c]{10}{*}{16} & \multirow[c]{5}{*}{KNN} & 1 & {\cellcolor[HTML]{FCB296}} \color[HTML]{000000} .686 & {\cellcolor[HTML]{FC9C7D}} \color[HTML]{000000} .617 & {\cellcolor[HTML]{FC8464}} \color[HTML]{F1F1F1} .535 & {\cellcolor[HTML]{FDCEBB}} \color[HTML]{000000} .789 & {\cellcolor[HTML]{FEE3D6}} \color[HTML]{000000} .867 & {\cellcolor[HTML]{FB7555}} \color[HTML]{F1F1F1} .485 & {\cellcolor[HTML]{FC9777}} \color[HTML]{000000} .599 & {\cellcolor[HTML]{FB6D4D}} \color[HTML]{F1F1F1} .463 & {\cellcolor[HTML]{FB7A5A}} \color[HTML]{F1F1F1} .502 & {\cellcolor[HTML]{FDC5AE}} \color[HTML]{000000} .752 & {\cellcolor[HTML]{FEE0D2}} \color[HTML]{000000} .848 & {\cellcolor[HTML]{FDD4C2}} \color[HTML]{000000} .808 & {\cellcolor[HTML]{FFF2EC}} \color[HTML]{000000} .968 \\
 &  & 2 & {\cellcolor[HTML]{FCB79C}} \color[HTML]{000000} .704 & {\cellcolor[HTML]{FCB296}} \color[HTML]{000000} .687 & {\cellcolor[HTML]{FC8E6E}} \color[HTML]{000000} .570 & {\cellcolor[HTML]{FDD2BF}} \color[HTML]{000000} .798 & {\cellcolor[HTML]{FEE3D6}} \color[HTML]{000000} .867 & {\cellcolor[HTML]{FB7252}} \color[HTML]{F1F1F1} .480 & {\cellcolor[HTML]{FC997A}} \color[HTML]{000000} .609 & {\cellcolor[HTML]{FB7757}} \color[HTML]{F1F1F1} .497 & {\cellcolor[HTML]{FC8B6B}} \color[HTML]{F1F1F1} .560 & {\cellcolor[HTML]{FDC5AE}} \color[HTML]{000000} .754 & {\cellcolor[HTML]{FEE0D2}} \color[HTML]{000000} .849 & {\cellcolor[HTML]{FDD5C4}} \color[HTML]{000000} .811 & {\cellcolor[HTML]{FFF3ED}} \color[HTML]{000000} .969 \\
 &  & 3 & {\cellcolor[HTML]{FCB79C}} \color[HTML]{000000} .705 & {\cellcolor[HTML]{FCB79C}} \color[HTML]{000000} .702 & {\cellcolor[HTML]{FC8B6B}} \color[HTML]{F1F1F1} .561 & {\cellcolor[HTML]{FDD2BF}} \color[HTML]{000000} .799 & {\cellcolor[HTML]{FEE3D6}} \color[HTML]{000000} .867 & {\cellcolor[HTML]{FB7555}} \color[HTML]{F1F1F1} .487 & {\cellcolor[HTML]{FC9B7C}} \color[HTML]{000000} .613 & {\cellcolor[HTML]{FB7757}} \color[HTML]{F1F1F1} .497 & {\cellcolor[HTML]{FC8B6B}} \color[HTML]{F1F1F1} .561 & {\cellcolor[HTML]{FDC6B0}} \color[HTML]{000000} .757 & {\cellcolor[HTML]{FEE0D2}} \color[HTML]{000000} .848 & {\cellcolor[HTML]{FDD3C1}} \color[HTML]{000000} .802 & {\cellcolor[HTML]{FFF3ED}} \color[HTML]{000000} .970 \\
 &  & 5 & {\cellcolor[HTML]{FCB69B}} \color[HTML]{000000} .699 & {\cellcolor[HTML]{FCB89E}} \color[HTML]{000000} .707 & {\cellcolor[HTML]{FC8E6E}} \color[HTML]{000000} .570 & {\cellcolor[HTML]{FDCCB8}} \color[HTML]{000000} .779 & {\cellcolor[HTML]{FEE3D6}} \color[HTML]{000000} .868 & {\cellcolor[HTML]{FB7151}} \color[HTML]{F1F1F1} .474 & {\cellcolor[HTML]{FC9B7C}} \color[HTML]{000000} .611 & {\cellcolor[HTML]{FB7C5C}} \color[HTML]{F1F1F1} .511 & {\cellcolor[HTML]{FC8969}} \color[HTML]{F1F1F1} .555 & {\cellcolor[HTML]{FCB99F}} \color[HTML]{000000} .711 & {\cellcolor[HTML]{FEE0D2}} \color[HTML]{000000} .848 & {\cellcolor[HTML]{FDCDB9}} \color[HTML]{000000} .781 & {\cellcolor[HTML]{FFF3ED}} \color[HTML]{000000} .969 \\
 &  & 10 & {\cellcolor[HTML]{FCB095}} \color[HTML]{000000} .681 & {\cellcolor[HTML]{FCB79C}} \color[HTML]{000000} .705 & {\cellcolor[HTML]{FC8767}} \color[HTML]{F1F1F1} .551 & {\cellcolor[HTML]{FCC2AA}} \color[HTML]{000000} .740 & {\cellcolor[HTML]{FEE3D6}} \color[HTML]{000000} .868 & {\cellcolor[HTML]{FA6849}} \color[HTML]{F1F1F1} .446 & {\cellcolor[HTML]{FC9070}} \color[HTML]{000000} .578 & {\cellcolor[HTML]{FB7858}} \color[HTML]{F1F1F1} .498 & {\cellcolor[HTML]{FB7D5D}} \color[HTML]{F1F1F1} .517 & {\cellcolor[HTML]{FCB99F}} \color[HTML]{000000} .711 & {\cellcolor[HTML]{FEE0D2}} \color[HTML]{000000} .848 & {\cellcolor[HTML]{FCC1A8}} \color[HTML]{000000} .739 & {\cellcolor[HTML]{FFF2EC}} \color[HTML]{000000} .967 \\
 & LinReg &  & {\cellcolor[HTML]{F03F2E}} \color[HTML]{F1F1F1} .329 & {\cellcolor[HTML]{F14331}} \color[HTML]{F1F1F1} .341 & {\cellcolor[HTML]{E43027}} \color[HTML]{F1F1F1} .279 & {\cellcolor[HTML]{E43027}} \color[HTML]{F1F1F1} .280 & {\cellcolor[HTML]{D42121}} \color[HTML]{F1F1F1} .221 & {\cellcolor[HTML]{ED392B}} \color[HTML]{F1F1F1} .311 & {\cellcolor[HTML]{F7593F}} \color[HTML]{F1F1F1} .402 & {\cellcolor[HTML]{F6583E}} \color[HTML]{F1F1F1} .398 & {\cellcolor[HTML]{F96044}} \color[HTML]{F1F1F1} .425 & {\cellcolor[HTML]{800610}} \color[HTML]{F1F1F1} -.026 & {\cellcolor[HTML]{FCB79C}} \color[HTML]{000000} .704 & {\cellcolor[HTML]{E43027}} \color[HTML]{F1F1F1} .280 & {\cellcolor[HTML]{F14130}} \color[HTML]{F1F1F1} .335 \\
 & \multirow[c]{4}{*}{LocalLinReg} & 2 & {\cellcolor[HTML]{FCB79C}} \color[HTML]{000000} .703 & {\cellcolor[HTML]{FCB296}} \color[HTML]{000000} .687 & {\cellcolor[HTML]{FC8969}} \color[HTML]{F1F1F1} .554 & {\cellcolor[HTML]{FDCEBB}} \color[HTML]{000000} .787 & {\cellcolor[HTML]{FEE3D6}} \color[HTML]{000000} .867 & {\cellcolor[HTML]{FB7353}} \color[HTML]{F1F1F1} .481 & {\cellcolor[HTML]{FC9D7F}} \color[HTML]{000000} .622 & {\cellcolor[HTML]{FB7656}} \color[HTML]{F1F1F1} .490 & {\cellcolor[HTML]{FC8969}} \color[HTML]{F1F1F1} .552 & {\cellcolor[HTML]{FDC7B2}} \color[HTML]{000000} .764 & {\cellcolor[HTML]{FEE0D2}} \color[HTML]{000000} .848 & {\cellcolor[HTML]{FDD4C2}} \color[HTML]{000000} .808 & {\cellcolor[HTML]{FFF3ED}} \color[HTML]{000000} .969 \\
 &  & 3 & {\cellcolor[HTML]{FCB79C}} \color[HTML]{000000} .704 & {\cellcolor[HTML]{FCB499}} \color[HTML]{000000} .695 & {\cellcolor[HTML]{FC8262}} \color[HTML]{F1F1F1} .533 & {\cellcolor[HTML]{FDCEBB}} \color[HTML]{000000} .785 & {\cellcolor[HTML]{FEE3D6}} \color[HTML]{000000} .867 & {\cellcolor[HTML]{FC8262}} \color[HTML]{F1F1F1} .531 & {\cellcolor[HTML]{FC9C7D}} \color[HTML]{000000} .615 & {\cellcolor[HTML]{FB7555}} \color[HTML]{F1F1F1} .489 & {\cellcolor[HTML]{FC8969}} \color[HTML]{F1F1F1} .553 & {\cellcolor[HTML]{FDC7B2}} \color[HTML]{000000} .763 & {\cellcolor[HTML]{FEE0D2}} \color[HTML]{000000} .848 & {\cellcolor[HTML]{FDD3C1}} \color[HTML]{000000} .805 & {\cellcolor[HTML]{FFF3ED}} \color[HTML]{000000} .969 \\
 &  & 5 & {\cellcolor[HTML]{FCAF93}} \color[HTML]{000000} .680 & {\cellcolor[HTML]{FB7353}} \color[HTML]{F1F1F1} .484 & {\cellcolor[HTML]{FC7F5F}} \color[HTML]{F1F1F1} .519 & {\cellcolor[HTML]{FED8C7}} \color[HTML]{000000} .821 & {\cellcolor[HTML]{FEE3D6}} \color[HTML]{000000} .868 & {\cellcolor[HTML]{FB7A5A}} \color[HTML]{F1F1F1} .503 & {\cellcolor[HTML]{FC8565}} \color[HTML]{F1F1F1} .542 & {\cellcolor[HTML]{FA6849}} \color[HTML]{F1F1F1} .446 & {\cellcolor[HTML]{FC8060}} \color[HTML]{F1F1F1} .525 & {\cellcolor[HTML]{FDCBB6}} \color[HTML]{000000} .774 & {\cellcolor[HTML]{FEE0D2}} \color[HTML]{000000} .849 & {\cellcolor[HTML]{FEE1D3}} \color[HTML]{000000} .852 & {\cellcolor[HTML]{FFF3ED}} \color[HTML]{000000} .972 \\
 &  & 10 & {\cellcolor[HTML]{FCAE92}} \color[HTML]{000000} .676 & {\cellcolor[HTML]{F6563D}} \color[HTML]{F1F1F1} .394 & {\cellcolor[HTML]{FC9070}} \color[HTML]{000000} .578 & {\cellcolor[HTML]{FEDCCD}} \color[HTML]{000000} .836 & {\cellcolor[HTML]{FEE3D6}} \color[HTML]{000000} .868 & {\cellcolor[HTML]{FC8161}} \color[HTML]{F1F1F1} .527 & {\cellcolor[HTML]{FB7A5A}} \color[HTML]{F1F1F1} .505 & {\cellcolor[HTML]{F96044}} \color[HTML]{F1F1F1} .424 & {\cellcolor[HTML]{FC7F5F}} \color[HTML]{F1F1F1} .519 & {\cellcolor[HTML]{FDCDB9}} \color[HTML]{000000} .784 & {\cellcolor[HTML]{FEE0D2}} \color[HTML]{000000} .848 & {\cellcolor[HTML]{FEE0D2}} \color[HTML]{000000} .850 & {\cellcolor[HTML]{FFF4EE}} \color[HTML]{000000} .974 \\

\end{tabular}